\documentclass[]{style}

\definecolor{lastauthor}{RGB}{143, 68, 115}

\title{From Masks to Pixels and Meaning:
A New Taxonomy, Benchmark, and Metrics for VLM Image Tampering} 

\author[1,2,\star]{{Xinyi Shang}}
\author[1,\star]{{Yi Tang}}
\author[1,\star]{{Jiacheng Cui}}
\author[1]{{Ahmed Elhagry}}
\author[1]{{Salwa K. Al Khatib}}
\author[1]{{Sondos Mahmoud Bsharat}}
\author[1]{{Jiacheng Liu}}
\author[1]{{Xiaohan Zhao}}
\author[2]{{Jing-Hao Xue}}
\author[1]{{Hao Li}}
\author[1]{{Salman Khan}}
\author[1,\dagger]{{Zhiqiang Shen}}

\affiliation[1]{Mohamed bin Zayed University of Artificial Intelligence}
\affiliation[2]{University College London}

\contribution[\star]{Equal Contribution}
\contribution[\dagger]{Corresponding author}

\usepackage[table]{xcolor} 
 
\newcommand{\myrowcolor}{\rowcolor[HTML]{E6F7FD}}
\abstract{
Existing tampering detection benchmarks largely rely on object masks, which severely misalign with the true edit signal: many pixels inside a mask are untouched or only trivially modified, while subtle yet consequential edits outside the mask are treated as natural. We reformulate VLM image tampering from coarse region labels to a pixel-grounded, meaning and language-aware task. First, we introduce a taxonomy spanning edit primitives (replace/remove/splice/inpaint/attribute/colorization, etc.) and their semantic class of tampered object, linking low-level changes to high-level understanding. Second, we release a new benchmark with per-pixel tamper maps and paired category supervision to evaluate detection and classification within a unified protocol. Third, we propose a training framework and evaluation metrics that quantify pixel-level correctness with localization to assess confidence or prediction on true edit intensity, and further measure tamper meaning understanding via semantics-aware classification and natural language descriptions for the predicted regions. We also re-evaluate the existing strong segmentation/localization baselines on recent strong tamper detectors and reveal substantial over- and under-scoring using mask-only metrics, and expose failure modes on micro-edits and off-mask changes. Our framework advances the field from masks to pixels, meanings and language descriptions, establishing a rigorous standard for tamper localization, semantic classification and description. Code and benchmark data are available at \url{https://github.com/VILA-Lab/PIXAR}.
}

\correspondence{Zhiqiang Shen (\email{Zhiqiang.Shen@mbzuai.ac.ae})}

\usepackage{arydshln}
\usepackage[table]{xcolor}  

\usepackage{epigraph}
\usepackage[normalem]{ulem}
\definecolor{quotemark}{gray}{0.7}
\makeatletter
\def\fquote{%
    \@ifnextchar[{\fquote@i}{\fquote@i[]}
           }%
\def\fquote@i[#1]{%
    \def\tempa{#1}%
    \@ifnextchar[{\fquote@ii}{\fquote@ii[]}
                 }%
\def\fquote@ii[#1]{%
    \def\tempb{#1}%
    \@ifnextchar[{\fquote@iii}{\fquote@iii[]}
                      }%
\def\fquote@iii[#1]{%
    \def\tempc{#1}%
    \vspace{1em}%
    \noindent%
    \begin{list}{}{%
         \setlength{\leftmargin}{0.05\textwidth}%
         \setlength{\rightmargin}{0.05\textwidth}%
                  }%
         \item[]%
         \begin{picture}(0,0)%
         \put(-8,-5){\makebox(0,0){\scalebox{2}{\textcolor{quotemark}{``}}}}%
         \end{picture}%
         \begingroup\itshape}%
 \def\endfquote{%
 \endgroup\par%
 \makebox[0pt][l]{%
 \hspace{0.27\textwidth}%
 \begin{picture}(0,0)(0,0)%
 \put(60,20){\makebox(0,0){%
 \scalebox{2}{\color{quotemark}''}}}%
 \end{picture}}%
 \ifx\tempa\empty%
 \else%
    \ifx\tempc\empty%
       \hfill\rule{110pt}{0.5pt}\\\mbox{}\hfill\tempa,\ \emph{\tempb}%
   \else%
       \hfill\rule{100pt}{0.5pt}\\\mbox{}\hfill\tempa,\ \emph{\tempb},\ \tempc%
   \fi\fi\par%
   \vspace{0.5em}%
 \end{list}%
 }%
 \makeatother

\newcommand{\algopt}{\textsc{\texttt{PIXAR}}\xspace}

\definecolor{orange}{RGB}{178,92,35}
\definecolor{green1}{RGB}{95,145,51}

\definecolor{red1}{RGB}{197,64,57}
\definecolor{blue1}{RGB}{59,130,220}
\definecolor{green2}{RGB}{82,181,150}
\definecolor{purple1}{RGB}{105,93,223}
\definecolor{orange1}{RGB}{164, 47, 54}
\definecolor{green3}{RGB}{94,145,51}

\definecolor{c_step1}{RGB}{60, 105, 199}
\definecolor{c_step2}{RGB}{159,37,28}
\definecolor{c_step3}{RGB}{107,40,157}

\providecommand{\xx}{\mathbf{x}}

\newenvironment{talign*}
{\csname align*\endcsname}
{\endalign}

\usepackage[utf8]{inputenc}         %
\usepackage[T1]{fontenc}            %
\usepackage{url}                    %
\usepackage{booktabs}               %
\usepackage{amsfonts}               %
\usepackage{nicefrac}               %
\usepackage{microtype}              %
\usepackage{algorithm}
\usepackage{algorithmic}
\usepackage{graphicx}
\usepackage{subcaption}
\usepackage[flushleft]{threeparttable}
\usepackage{float}
\usepackage{multirow}
\usepackage{xspace}
\usepackage{enumitem}
\usepackage[font=small]{caption}
\usepackage{autobreak}
\usepackage{sidecap}
\usepackage{wrapfig}
\usepackage{bbding}
\usepackage[toc, page, header]{appendix}
\usepackage{tikz}
\usepackage{pifont}
\usepackage{mdframed}
\usepackage{colortbl}

\definecolor{coral}{RGB}{255,127,80}
\definecolor{darkgreen}{RGB}{0,100,0}
\definecolor{darkyellow}{RGB}{204,153,0}
\definecolor{salmon}{RGB}{250,128,114}
\definecolor{darkred}{RGB}{150,0,0}
\newcommand{\darkredtext}[1]{{\color{darkred}#1}}

\newcommand{\secref}[1]{\hyperref[#1]{\darkredtext{Sec.~\ref*{#1}}}}
\newcommand{\thmref}[1]{\hyperref[#1]{\darkredtext{Thm.~\ref*{#1}}}}
\newcommand{\defref}[1]{\hyperref[#1]{\darkredtext{Def.~\ref*{#1}}}}
\newcommand{\propref}[1]{\hyperref[#1]{\darkredtext{Prop.~\ref*{#1}}}}
\newcommand{\assumpref}[1]{\hyperref[#1]{\darkredtext{Assump.~\ref*{#1}}}}
\newcommand{\remarkref}[1]{\hyperref[#1]{\darkredtext{Rem.~\ref*{#1}}}}
\newcommand{\hypref}[1]{\hyperref[#1]{\darkredtext{Hyp.~\ref*{#1}}}}
\newcommand{\conjref}[1]{\hyperref[#1]{\darkredtext{Conj.~\ref*{#1}}}}
\newcommand{\lemref}[1]{\hyperref[#1]{\darkredtext{Lem.~\ref*{#1}}}}
\newcommand{\corref}[1]{\hyperref[#1]{\darkredtext{Cor.~\ref*{#1}}}}
\newcommand{\noteref}[1]{\hyperref[#1]{\darkredtext{Nota.~\ref*{#1}}}}
\newcommand{\claimref}[1]{\hyperref[#1]{\darkredtext{Clm.~\ref*{#1}}}}
\newcommand{\obsref}[1]{\hyperref[#1]{\darkredtext{Obs.~\ref*{#1}}}}
\newcommand{\algref}[1]{\hyperref[#1]{\darkredtext{Alg.~\ref*{#1}}}}
\newcommand{\figref}[1]{\hyperref[#1]{\darkredtext{Fig.~\ref*{#1}}}}
\newcommand{\tabref}[1]{\hyperref[#1]{\darkredtext{Tab.~\ref*{#1}}}}
\newcommand{\appref}[1]{\hyperref[#1]{\darkredtext{App.~\ref*{#1}}}}
\renewcommand{\eqref}[1]{\hyperref[#1]{\darkredtext{Eq.~\ref*{#1}}}}

\begin{document}

\maketitle

\section{Introduction}
\label{sec:intro}

\begin{fquote}[Morpheus][The Matrix (1999)]{What is real? How do you define `real'?}
\end{fquote}

Advances in generative AI~\citep{wu2025qwen,comanici2025gemini,xia2024gsva} have enabled the creation of photorealistic imagery, posing serious threats to digital media authenticity and trust~\citep{xu2023combating,pal2024semi,monteith2024artificial}.
Among these manipulations, fine-grained tampering~\citep{gupta2025multiverse,guo2023hierarchical,nandi2023trainfors} is particularly insidious, as it subtly modifies partial regions of real images while remaining imperceptible to both human and conventional forensic methods~\citep{guo2025language,huang2025sida}.
Consequently, developing robust detectors for fine-grained tampering is both a critical research challenge and a societal necessity.

\begin{figure}[!t]
    \centering
    \includegraphics[width=.85\linewidth]{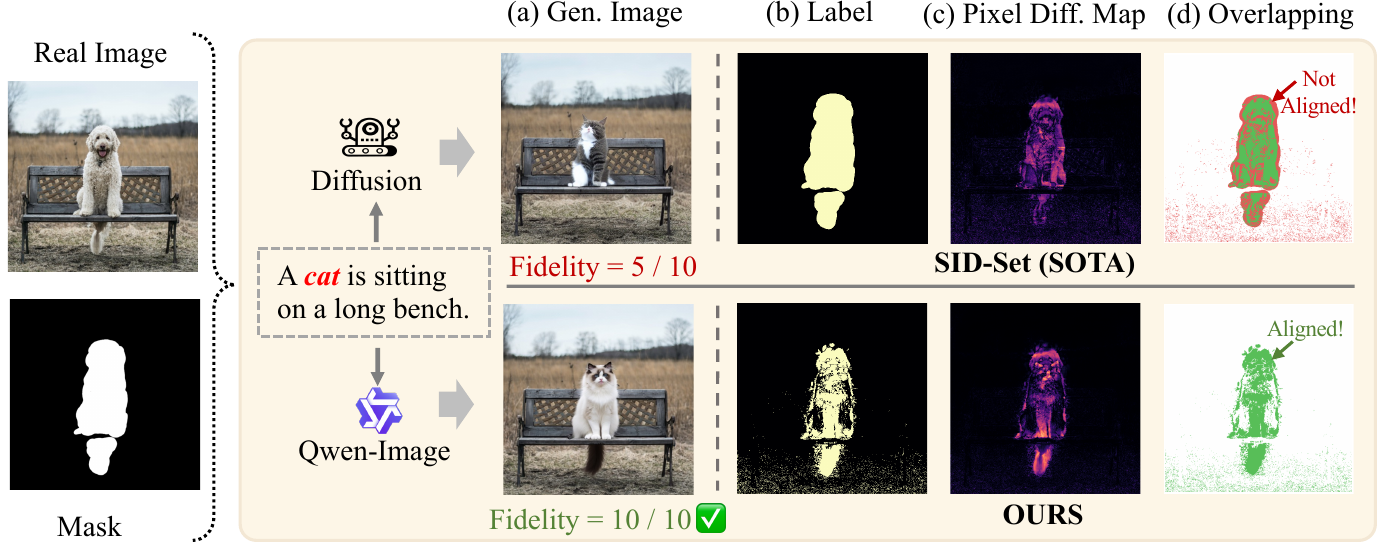}
   \caption{\textbf{Pitfalls of Current Benchmarks and Our Remedy.}
            (a) They contain unrealistic samples.
            (b)–(d) Their widely adopted mask-based label contains large regions of \textcolor{red1}{not aligned} pixels with the true generative pixels.
             In contrast, our pixel label is precisely \textcolor{green1}{aligned} with the true generative pixels.  
            }
    \label{fig:motivation}
    \vspace{-10pt}
\end{figure}

Yet most established benchmarks still annotate ``where the edit is'' using coarse object masks, as summarized in \tabref{tab:summary_datasets}. This practice implicitly assumes that edits are spatially confined to a pre-specified region and that all pixels within that region are equally manipulated. In practice, however, the edit signal is neither spatially nor metrically uniform: many pixels inside a mask remain unchanged or only trivially perturbed, while visually consequential adjustments (e.g., relighting halos, color bleeding, deblurring, seam removal) frequently extend outside the mask. As a result, mask-only evaluation conflates unedited pixels with true tamper evidence and ignores off-mask artifacts, distorting both detector training and measurement.

We make this pixel-level real tamper explicit by contrasting per-pixel difference maps between original and tampered images (as shown in \figref{fig:motivation} (c)), and visualize prior weaker solution SID-Set~\citep{huang2025sida}'s mask-defined labels in \figref{fig:motivation} (b). The visualization of \figref{fig:motivation} (d) reveals widespread misalignment (highlighted as \textcolor{red1}{red points}) between the human-defined groundtruth and true pixel-level tamper, i.e., untouched pixels falsely labeled as ``tampered'' inside the mask and edited pixels incorrectly treated as ``real'' outside the mask. These label errors significantly penalize models for detecting genuine generative artifacts that fall beyond the mask boundary and, conversely, reward models that overfit to coarse shapes rather than the true edit footprint. Our analysis challenges the prevailing assumption that generative pipelines modify all masked pixels while preserving all unmasked areas, raising a fundamental question for the generative era: where, precisely, is the boundary between the ``real'' and the ``generated'', and how should benchmarks encode that boundary?

\begin{figure}[!t]
    \centering
    \includegraphics[width=.85\linewidth]{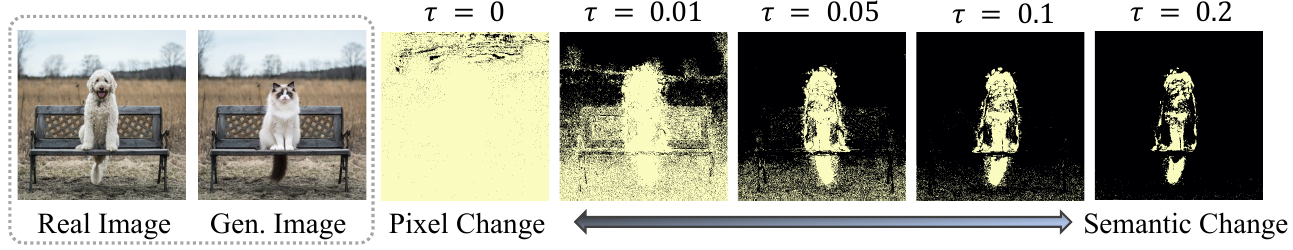}
   \caption{Visualization of our pixel-level label under different $\tau$.
   Small $\tau$ emphasizes sensitivity to pixel edits, while larger $\tau$ emphasizes semantic changes.
   }
    \label{fig:vis_tau}
\end{figure}

To address this, we reformulate VLM image tampering as a \uline{\bf pix}el-grounded, meaning and text-\uline{\bf a}wa\uline{\bf r}e task and construct a new benchmark called \algopt. Concretely, we derive a difference map between the original and edited images and convert it into a binary supervision signal via a tunable threshold $\tau$. The resulting label map $\mathbf{M}_{\tau}$ captures the spatial support of the edit at a controllable intensity level (\figref{fig:vis_tau}): small $\tau$ emphasizes sensitivity to micro-edits, while larger $\tau$ emphasizes conservative, high-confidence changes. This thresholded construction decouples where an edit occurs (localization) from how strongly it manifests (intensity), enabling principled sweeps over $\tau$ to select operating points that best correlate with human judgments and downstream scenario use cases. Consequently, our formulation aligns evaluation with the physical tampering signal rather than proxy geometries.

Building on this reformulation, we introduce a large-scale benchmark -- over $380\text{K}$ carefully curated training image pairs with rich, standardized metadata, and a well-balanced test set containing $40\text{K}$ image pairs with pixel-level and semantic-level labels.
Each pair comprises a real source image, its tampered counterpart, the recommended binary pixel label $\mathbf{M}_{\tau}$ from a default $\tau$, and the raw per-pixel difference map from which alternative labels for other $\tau$ values can be derived.
To ensure manipulation diversity, our pipeline integrates eight editing strategies instantiated via state-of-the-art both open- and closed-source generative models, including Flux.2~\citep{flux2}, Gemini 2.5~\citep{gemini25}, Gemini 3~\citep{gemini3}, GPT-image-1.5~\citep{gptimage15}, Qwen-Image~\citep{wu2025qwen}, Seedream 4.5~\citep{seedream2025seedream40}.
The strategies span replace/remove/splice/inpaint/attribute/colorization and related primitives, and we manually annotate the semantic class of the tampered target to connect low-level footprints to high-level semantics.
Finally, we design a rigorous multi-stage filtering pipeline to guarantee the fidelity of tampered images and the precision of the corresponding labels.
These designs yield a dataset that is simultaneously pixel-faithful and semantically structured, supporting detection, localization, and semantic understanding within a single protocol.

In summary, our contributions are threefold:
\begin{enumerate}[itemsep=3pt, topsep=3pt]
\item We are the \textit{first} to expose the core flaw of mask-based benchmarks, i.e. {\em misalignment with true generative edits}, and redefine tampering with pixel-level and semantic supervision to yield precise, faithful labels.
\item We build \algopt, a large-scale, high-fidelity benchmark using state-of-the-art VLMs and human semantic label annotation, integrating 8 manipulation types with rigorous effectiveness and fidelity checks and accurate pixel annotations, establishing a principled foundation for the generative era.
\item We introduce a training framework and pixel-wise, realistic evaluation metrics, and extensively re-evaluate the state-of-the-art detectors on our \algopt, revealing key limitations (e.g., micro- and off-mask failures) and setting stronger, more reliable baselines.
\end{enumerate}

\begin{table}[b]
    \centering
    \vspace{-0.1in}
    \caption{Details of recent publicly available tampering benchmarks.
    The \textbf{Task} column indicates the evaluation type, where ``B-Cla.'' refers to binary classification (
    ``real'' vs. ``tampered''), and ``T-Loc.'' denotes tampering localization.
    ``Multi-Object'' means multiple objects within one image can be tampered with.
    }
    \label{tab:summary_datasets}
    \resizebox{.9\textwidth}{!}{
    \begin{tabular}{lccccc}
    \toprule
    \textbf{Dataset} & \textbf{Year} & \textbf{Task} & \textbf{Multi-Object} & \textbf{Fidelity Check} & \textbf{Ground Truth} \\
    \midrule
    ArtiFact~\citep{rahman2023artifact} & 2023 & B-Cla. & \textcolor{red1}{\ding{55}} & \textcolor{red1}{\ding{55}} & -\\
    TrainFors~\citep{nandi2023trainfors} & 2024 & B-Cla. \& T-Loc. & \textcolor{red1}{\ding{55}} & \textcolor{red1}{\ding{55}} & Mask \\
    SIDBench~\citep{schinas2024sidbench} & 2024 & B-Cla. & \textcolor{red1}{\ding{55}} & \textcolor{red1}{\ding{55}} & - \\
    DiFF~\citep{cheng2024diffusion} & 2024 & B-Cla. & \textcolor{red1}{\ding{55}} & \textcolor{red1}{\ding{55}} & - \\
    M3Dsynth~\citep{zingarini2024m3dsynth} & 2024 & B-Cla. \& T-Loc. & \textcolor{red1}{\ding{55}} & \textcolor{red1}{\ding{55}} & Mask \\
    SemiTruths~\citep{pal2024semi} & 2024 & B-Cla. \& T-Loc.  & \textcolor{red1}{\ding{55}} & \textcolor{red1}{\ding{55}} & Mask \\
    AI-Face~\citep{lin2025ai} & 2025 & B-Cla. & \textcolor{red1}{\ding{55}} & \textcolor{red1}{\ding{55}} & - \\
    SID-Set~\citep{huang2025sida} & 2025 & B-Cla. \& T-Loc. & \textcolor{red1}{\ding{55}} & \textcolor{red1}{\ding{55}} & Mask  \\
    \myrowcolor
    \textbf{\algopt} & \textbf{2026} & B-Cla. \& T-Loc. & \textcolor{green1}{\ding{51}} & \textcolor{green1}{\ding{51}} & Pixel \& Semantics \\
    \bottomrule
    \end{tabular}}
\end{table}

\section{Related Work}
\label{sec:related_work}
\noindent{\bf Tampered Image Datasets.}
The rapid evolution of generative models, from Generative Adversarial Networks (GANs)~\citep{zhu2017unpaired,abdal2019image2stylegan,xia2022gan} to diffusion models~\citep{nichol2021glide,croitoru2023diffusion,yang2023diffusion,rombach2022high} and large Vision-Language Models (VLMs)~\citep{zhang2024vision,comanici2025gemini,wu2025qwen}, has necessitated the parallel development of robust and reliable detection benchmarks.
Early benchmarks primarily focus on full-image generation (e.g., text-to-image), training detectors for binary real-versus-fake classification~\citep{zhu2023genimage,zhong2023rich,lu2023seeing}.
However, as generative manipulations become increasingly subtle, recent research has shifted toward fine-grained tampering detection~\citep{huang2025sida}.
Recently, SID-Set~\citet{huang2025sida} employs Stable Diffusion–based inpainting~\citep{rombach2022high} to construct a benchmark for tampering localization in social media images.
Despite this progress, as summarized in \tabref{tab:summary_datasets}, existing datasets largely rely on object masks as ground truth, leading to substantial misalignment with the true edit signal and fundamentally impairing detectors from learning true tampering footprints.
In contrast, we redefine tampering with pixels, meanings, and language descriptions to yield precise, faithful supervision.

\noindent{\bf Tampered Image Detection.}
Tampering detection is commonly formulated as a classification task using CNN- or Transformer-based architectures~\citep{Capsule-Forensics,chen2022self}.
In addition, several studies~\citet{tan2024frequency,jeong2022frepgan} explore the frequency domain to capture generation-specific artifacts, while reconstruction-based methods such as RECCE~\citep{RECCE} aim to improve feature robustness through reconstruction learning.
Although these models achieve strong performance on images produced by seen generative models, their generalization to unseen ones remains limited.
To mitigate this issue, CNNSpot~\citet{CNNDetection} learns universal CNN artifacts, Fusing~\citet{fusing} combines global and local representations through attention-based feature fusion, UnivFD~\citet{UnivFD} leverages the CLIP feature space for training-free real-fake discrimination, and LGrad~\citet{LGrad} maps images into gradient space to achieve model-driven generalization.
More recently, Vision-Language Models have been introduced for tampering detection, such as SIDA~\citep{huang2025sida}, which fine-tunes LLaVA-based multimodal models, AntifakePrompt~\citep{chang2023antifakeprompt}, which employs prompt tuning to improve detection accuracy and cross-model generalization, and FakeShield~\citep{xu2024fakeshield}, which leverages VLM and a decoupled architecture to provide explainable detection and precise localization across diverse forgery domains, supported by the multi-modal MMTD-Set.

\begin{figure}[!t]
    \centering
    \includegraphics[width=\linewidth]{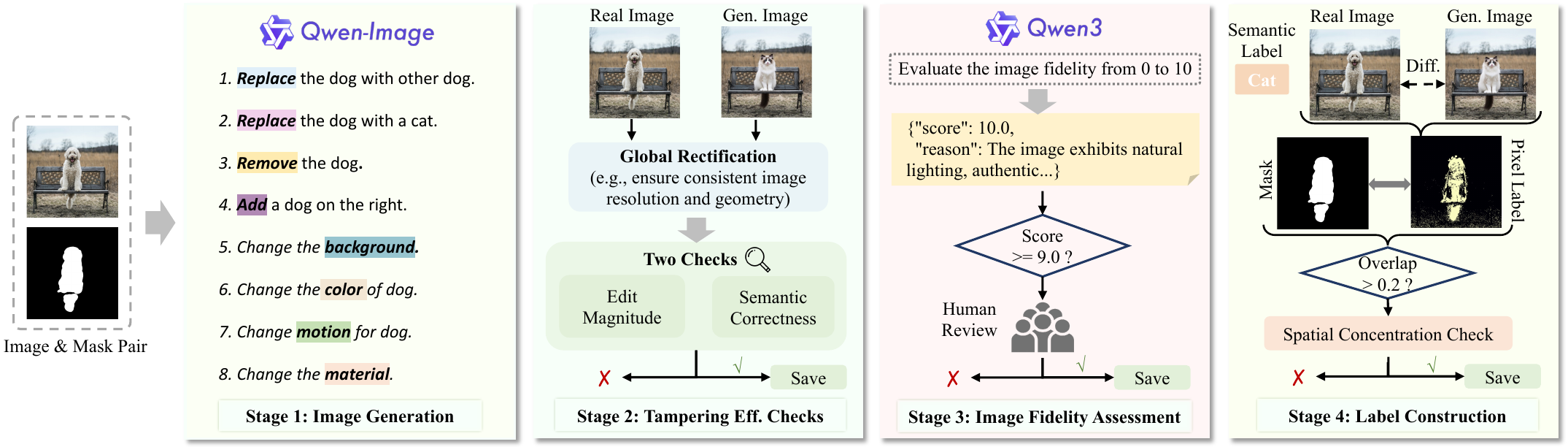}
    \caption{\textbf{Overview of our \algopt generation pipeline.}
    The pipeline consists of four stages.
    \textit{Stage 1: Image Generation:} We generate images with 8 tampering types.
    \textit{Stage 2: Tampering Effectiveness Checks:} We filter out ineffective tampered images.
    \textit{Stage 3: Image Fidelity Assessment:} The generated images are first assessed by Qwen3~\citep{yang2025qwen3} and then reviewed by human annotators.
    \textit{Stage 4: Label Construction:} We manually annotate semantic labels and construct faithful pixel-level annotations.}
    \label{fig:generation_pipeline}
    \vspace{-10pt}
\end{figure}

\section{Benchmark Construction}
\label{sec:benchamrk}
To thoroughly train and evaluate image tampering detectors, we build \algopt, a large-scale benchmark over $380\text{K}$ carefully curated training image pairs with rich, standardized metadata, and a well-balanced test set containing $40\text{K}$ image pairs with pixel-level and semantic-level labels.
The design of \algopt is guided by three principles: (i) \textbf{\textcolor{blue1}{Diversity:}} incorporating 8 tampering types that align well with real-world scenarios and demands;
(ii) \textbf{\textcolor{orange}{Fidelity:}} implementing rigorous fidelity checks to filter out low-fidelity samples;
and (iii) \textbf{\textcolor{green1}{Precision:}} ensuring precise labels for true tampering.
Accordingly, we propose a four-stage generation pipeline, illustrated in~\figref{fig:generation_pipeline}.
The following subsections detail each stage.
Additional implementations and the construction of balanced test data are provided in ~\appref{sec:app_benchamrk} and ~\appref{sec:app_test_data}, respectively.

\subsection{Image Generation}

\noindent{\bf Data Source and Generative Models.}
We use real source images from the COCO~\citep{lin2014microsoft}, a large-scale benchmark with diverse scenes, objects, and contexts\footnote{The project is stage-wise, and the images are continually expanded with more sources in our further versions.}.
For training set, we employ Qwen-Image VLMs~\citep{wu2025qwen} due to their significant advances in complex text rendering and precise image editing.
For test set, we use both open- and closed-source generative models, including Flux.2~\citep{flux2}, Gemini 2.5~\citep{gemini25}, Gemini 3~\citep{gemini3}, GPT-image-1.5~\citep{gptimage15}, Qwen-Image~\citep{wu2025qwen}, Seedream 4.5~\citep{seedream2025seedream40}.

\begin{figure*}[!t]
    \centering
    \includegraphics[width=\linewidth]{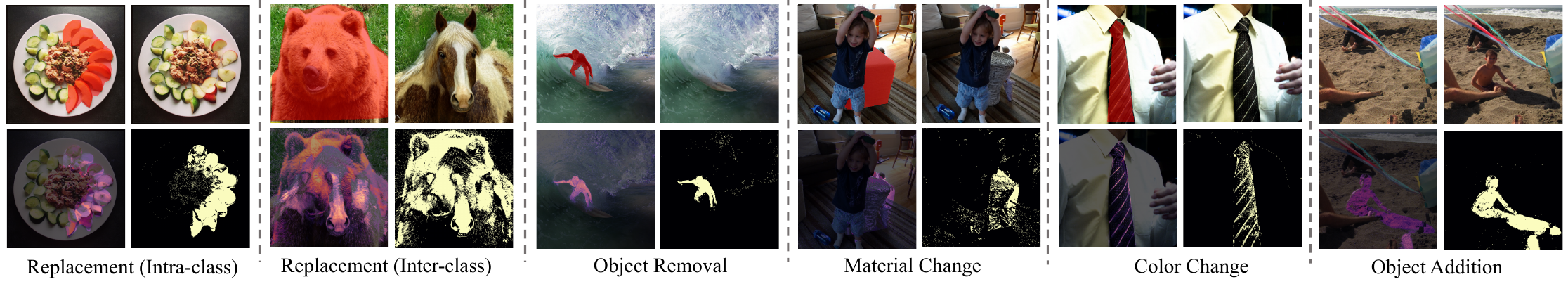}
    \vspace{-10pt}
    \caption{\textbf{Visualization of various tampering types in \algopt.}
    For each type, from top-left to bottom-right, the images show: original image with red shading indicating the modified object, tampered image, pixel-difference map overlaid on the tampered image, and our pixel-level label.
    }
    \label{fig:vis_benchmark}
    \vspace{-5pt}
\end{figure*}

\vspace{3pt}
\noindent{\bf Diverse and Practical Tampering Types.}
\label{sec:tamper_framework}
Existing benchmarks mainly employ inter-class replacement for tampered image generation~\citep {huang2025sida,zingarini2024m3dsynth,nandi2023trainfors}, which fails to reflect the complexity of real-world manipulations.
To align well with real-world tampered scenarios and practical demands, we first analyzed large-scale Internet images to define $8$ manipulation types (see Stage~1 of~\figref{fig:generation_pipeline}).
Visual examples are shown in~\figref{fig:vis_benchmark}, and more details are provided in \appref{app_sec:stage_1}.

\begin{wrapfigure}{r}{0.48\linewidth} 
    \vspace{-15pt}
    \centering
    \begin{subfigure}[b]{0.48\linewidth}
        \centering
        \includegraphics[width=\linewidth]{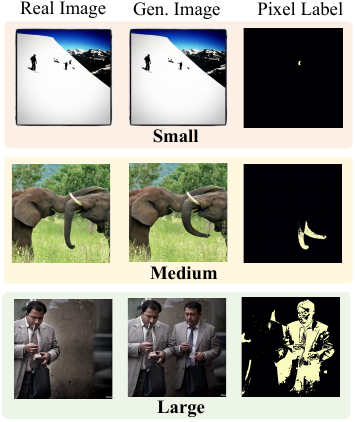}
        \caption{Sizes.}
        \label{fig:app_vis_size}
    \end{subfigure} 
    \begin{subfigure}[b]{0.48\linewidth}
        \centering
        \includegraphics[width=\linewidth]{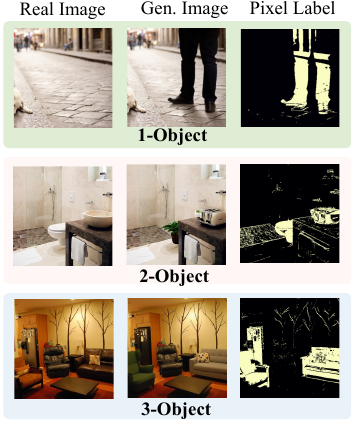}
        \caption{Complexities.}
        \label{fig:app_vis_multiple}
    \end{subfigure}
    \vspace{-5pt}
    \caption{Visualization of tampered images with varying (a) tampered sizes and (b) tampered complexities.}
    \vspace{-10pt}
    \label{fig:data_vis}
\end{wrapfigure}

\vspace{3pt}
\noindent{\bf Diverse Tampered Sizes and Complexities.}
To rigorously evaluate the robustness and discriminative performance of detection models across a spectrum of difficulty levels, we control two critical factors: {\it tampered size} and {\it tampered complexity}.
Tampered size measures the extent of pixel-level modification, where smaller edits leave subtler traces and are thus harder to detect, while larger edits yield more significant artifacts.
Tampered complexity reflects the compositional structure of manipulations.
Beyond the conventional single-object edits (\tabref{tab:summary_datasets}), we apply a multi-object, sequential protocol to better reflect iterative real-world forgeries.
Visualizations and more implementation details are provided in \appref{app_sec:stage_1}.

\subsection{Tampering Effectiveness Checks}
\label{sec:tampering_effectiveness_check}

Generative models frequently exhibit failure modes, such as unintended global repainting or trivial perturbations.
Therefore, we implement a rigorous filtering pipeline (illustrated in Stage 2 of \figref{fig:generation_pipeline}) to remove ineffective tampered images.
The pipeline consists of two sequential steps: (1) {\em Global Rectification}, which enforces pixel-space consistency between the generated image and the original image, and (2) {\em Edit Magnitude and Semantic Correctness Checks}, which verify whether the resulting perturbations are both meaningful and aligned with the intended semantics.

\vspace{3pt}
\noindent{\bf Global Rectification.}
In practice, many contemporary generative models (e.g., Gemini) do not support a given target resolution, which may result in pixel-space misalignment between the generated image and the original image.
Such geometric misalignment makes pixel-level difference maps unreliable, thereby corrupting the derived pixel labels $\mathbf{M}_{\tau}$.
To address this issue, we perform a geometric rectification step: we align $I_{\text{gen}}$ to $I_{\text{orig}}$ via feature matching, estimate a homography with RANSAC~\citep{fischler1981random}, and then recompute the difference map within this aligned coordinate pixel space.
The qualitative comparison before and after global rectification is provided in \figref{fig:app_Tampering_Effectiveness} (a), verifying the efficacy of our rectification.

\vspace{3pt}
\noindent{\bf Edit Magnitude and Semantic Correctness Checks.}
After rectification, we rigorously evaluate whether the generative model successfully executed the requested edit for the original image.
We observe three common failure cases: (i) \textit{near-zero tampering}, where $I_{\text{gen}}$ is almost identical to $I_{\text{orig}}$, yielding negligible signal in $\mathbf{M}_{\tau}$; (ii) \textit{unintended global editing}, where large image regions are repainted beyond the target area, producing extensive noise in $\mathbf{M}_{\tau}$; and
(iii) \textit{unintended semantic edits}, where the visual change does not correspond to the text instruction.
Accordingly, we assess validity from:
(a) edit magnitude, to exclude trivial or overly global changes, and
(b) semantic correctness, to ensure the visual manipulation matches the instruction.
Visualization examples and implementation details are provided in \appref{app_sec:stage_2}.

\subsection{Image Fidelity Assessment}
\label{sec:image_fidelity_check}

\begin{wrapfigure}{r}{0.33\linewidth} 
    \vspace{-15pt}
    \centering
    \includegraphics[width=\linewidth]{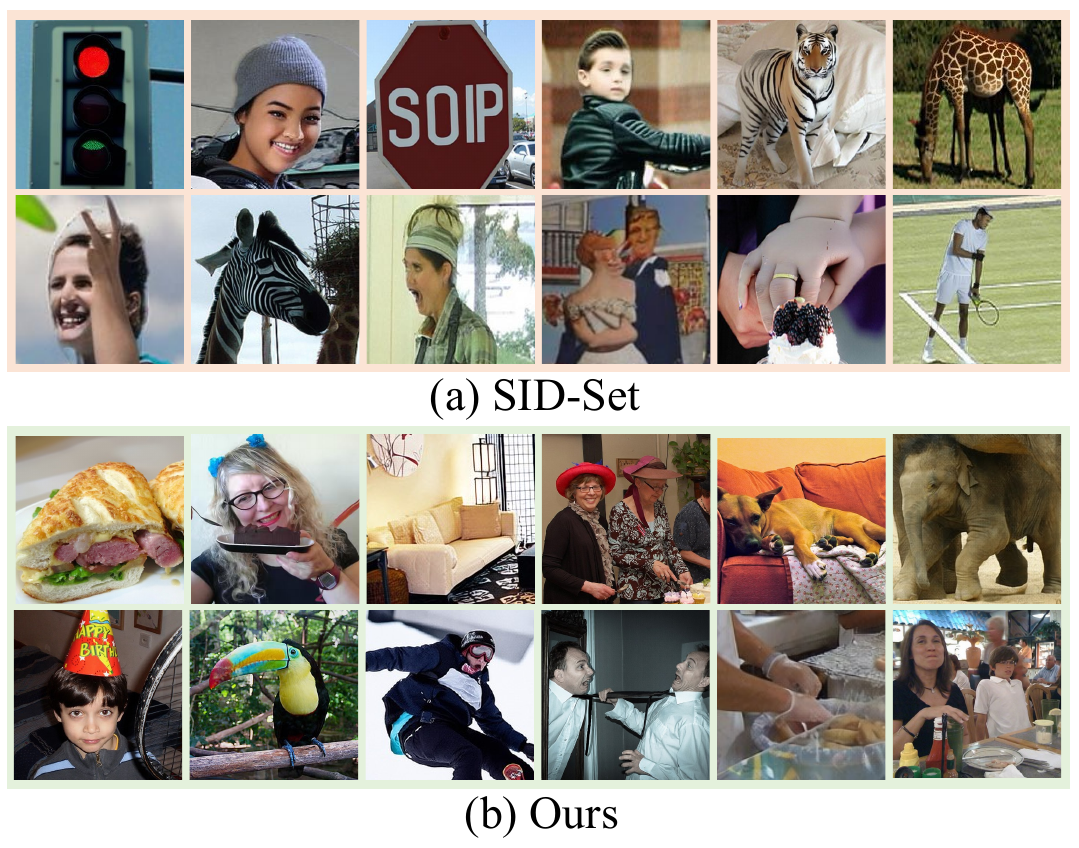}
    \vspace{-15pt}
    \caption{Comparison of fidelity between SID-Set and our \algopt.}
    \label{fig:compare_sida}
    \vspace{-15pt}
\end{wrapfigure}

The reliability of the benchmark critically depends on the fidelity of its samples.
Low-quality samples provide limited information for model training and evaluation~\citep{Zhang_2024_CVPR,gupta2025improving}.
To ensure high fidelity of images, we design a rigorous image fidelity assessment process that combines automated evaluation by the state-of-the-art VLM Qwen3~\citep{yang2025qwen3} with human expert evaluation, as illustrated in Stage 3 of \figref{fig:generation_pipeline}.

\vspace{3pt}
\noindent{\bf Automated VLM Assessment.}
Each tampered image candidate is first evaluated by Qwen3, which assigns a fidelity score from 0 to 10.
Only tampered images achieving a score of $\ge$ 9 are shortlisted for subsequent human evaluation.
This automated stage efficiently and effectively filters out low-fidelity images and thus improves the quality of our \algopt.

\vspace{3pt}
\noindent{\bf Human Expert Review.}
Furthermore, we employ 10 human experts to manually review the generated images and remove those that appear unrealistic.
Only samples that received a realism score of at least 4 out of 5 were retained.
This human-in-the-loop evaluation ensures that subtle inconsistencies or semantic anomalies undetectable by automated models are effectively filtered out.
We observe consistently high pass rates (90\%) for intra-class replacement, splicing, inpainting, attribute modification, and colorization.
In contrast, the inter-class replacement variant achieves a moderate pass rate of approximately 70\%, while the removal type shows the lowest pass rate at around 55\%.
Representative examples of filtered and retained samples are provided in \figref{fig:app_vis_human_filtering}.

After the stringent image filtering process, we further validate quality by randomly sampling tampered images from SID-Set~\citep{huang2025sida} and from our \algopt for comparison, as shown in \figref{fig:compare_sida}. Samples from SID-Set often appear visually unrealistic. In contrast, our dataset exhibits substantially higher fidelity, demonstrating a more reliable and principled foundation for tampered image detection.

\subsection{Label Construction}
To remedy the flaw of the existing mask-based label, we redefine tampering with pixel-level and semantic supervision.
Consequently, detectors trained on our \algopt learn not only \textit{where} image tampering occurs (local pixel detection) but also \textit{what} the tampered content means (semantic understanding).

\vspace{3pt}
\noindent{\bf Pixel Label.}
Given a pair of real and tampered images $(I_\text{orig}, I_\text{gen})$, we compute a difference map $\mathbf{D}$ that quantifies the absolute pixel-wise discrepancy:
\begin{equation}
\label{eq:soft_label}
    \setlength{\abovedisplayskip}{4pt}
    \setlength{\belowdisplayskip}{4pt}
    \mathbf{D}(\xx, y) = | I_\text{orig}(\xx, y) - I_\text{gen}(\xx, y) |,
\end{equation}
where $(\xx, y)$ indexes pixel coordinates.
We then obtain a binary supervision mask $\mathbf{M}_{\tau}$ by thresholding $\mathbf{D}$ with a tunable parameter $\tau$:
\begin{equation}
\label{eq:hard_label}
    \setlength{\abovedisplayskip}{4pt}
    \setlength{\belowdisplayskip}{4pt}
\mathbf{M}_{\tau}(\xx, y) = \mathbb{I}\big(\mathbf{D}(\xx, y) > \tau\big),
\end{equation}
where $\mathbb{I}(\cdot)$ denotes the indicator function.
The resulting $\mathbf{M}_{\tau}$ captures the spatial support of the edit at a controllable intensity level: a small $\tau$ emphasizes sensitivity to micro-edits, while a larger $\tau$ emphasizes high-confidence modifications, as illustrated in~\figref{fig:vis_tau}.
We provide more visualizations in \figref{fig:app_vis_tau} and a detailed analysis regarding the $\tau$ selection in~\secref{sec:exp_ablation}.

\vspace{3pt}
\noindent{\bf Semantic Label.}
For each tampered object, we manually annotate its corresponding semantic class label (e.g., ``cat'' for the tampered image) and record this information as metadata associated with the image.
Formally, let $\mathcal{C}$ be the set of object classes that may be tampered, with multi-label ground truth $
\mathbf{y} \in\{0,1\}^{|\mathcal{C}|}$, where $y_c=1$ if class $c$ contains a tampered instance.

\vspace{3pt}
\noindent{\bf Label Reliability Checks.}
Finally, we validate the reliability of the pixel-level label with respect to semantic supervision.
In practice, low-quality pixel labels arise from two issues: (i) \emph{pixel--semantic inconsistency}, where the pixel change fails to reflect a valid semantic edit; and (ii) \emph{poor spatial structure}, where the pixel label is semantically consistent yet overly dispersed (often dominated by background pixels), making it unreliable for supervision.
We address these failures using two checks:

\begin{wrapfigure}{r}{0.45\linewidth} 
    \centering
    \vspace{-15pt}
     \includegraphics[width=\linewidth]{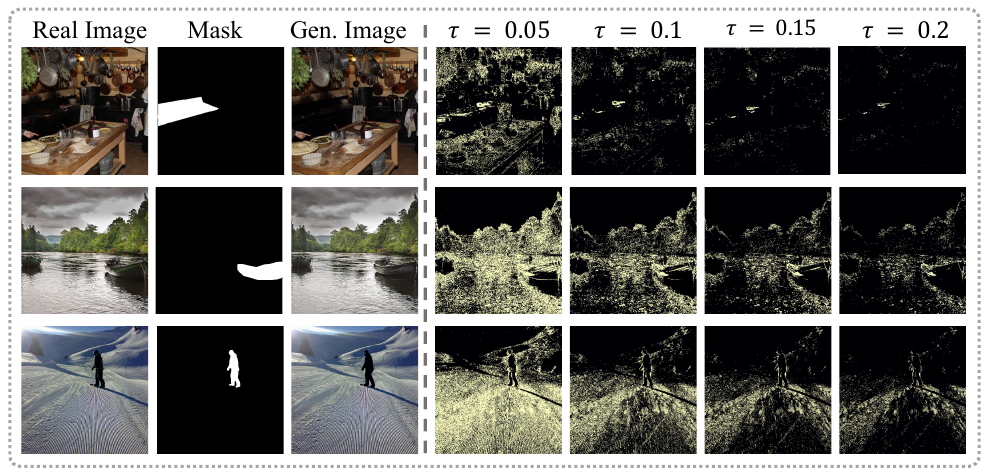}
     \vspace{-15pt}
    \caption{Examples of pixel--semantic inconsistency.}
    \label{fig:failure_gt}
    \vspace{-15pt}
\end{wrapfigure}

\emph{(1) Pixel--semantic consistency.}
As shown in~\figref{fig:failure_gt}, in some cases, replacing an object (e.g., the black oven in the first row) with a visually similar one may yield negligible $L_1$ differences due to near-identical color and texture.
Consequently, the pixel-level label underestimates the true semantic change, causing a pixel--semantic mismatch.
To ensure faithful pixel supervision, we compute the overlap ratio between tampered pixels and the input mask, and discard samples with overlap $<0.2$.
More examples and a detailed discussion of this threshold are provided in \appref{sec:app_stage_4}.

\emph{(2) Spatial concentration.}
Even when the semantic edit is correct, generation artifacts may introduce widespread background speckles, producing pixel labels that are scattered rather than object-shaped.
Examples are visualized in \figref{fig:app_Tampering_Effectiveness} (c).
Such dispersed labels are semantically consistent but structurally uninformative.
To remove them, we quantify the spatial concentration of $\mathbf{M}_{\tau}$ using: (i) a grid-based concentration ratio, defined as the fraction of grid cells required to cover 80\% of tampered pixels; and (ii) a local density score, defined as the median density of tampered pixels within a small neighborhood.
Based on these metrics, we discard \emph{Diverse} samples and retain only \emph{Concentrated} pixel labels that provide clean, structured supervision.
By filtering both inconsistent and dispersed cases, we obtain pixel supervision that is faithful to the edit and spatially well-formed.

\subsection{Metadata}
\begin{wrapfigure}{r}{0.5\linewidth} 
    \vspace{-15pt}
    \centering
    \includegraphics[width=\linewidth]{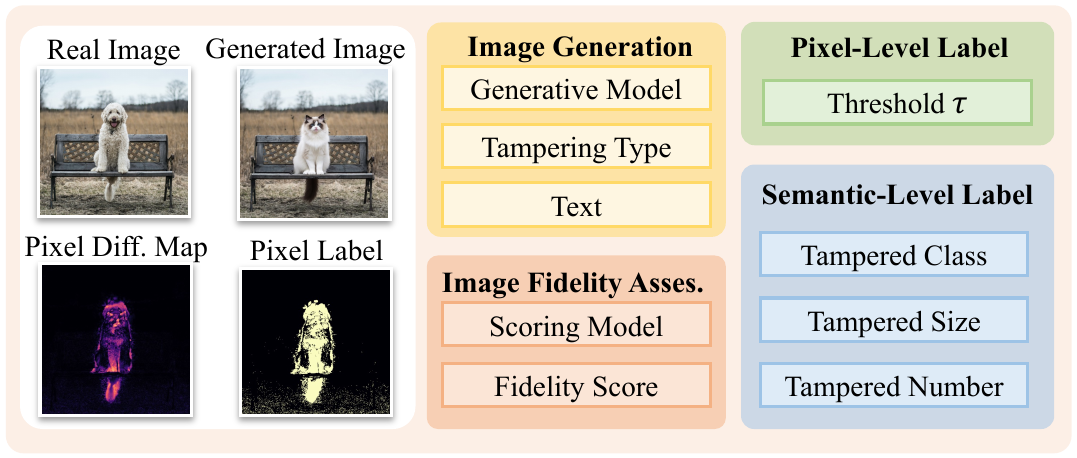}
    \vspace{-0.22in}
    \caption{\textbf{Composition of \algopt.}
    Each entry includes four images with detailed metadata.}
    \label{fig:metadata}
    \vspace{-10pt}
\end{wrapfigure}

Each entry in our \algopt\ benchmark encompasses four images accompanied by rich, standard metadata, providing comprehensive information about every stage of the tampering process, as illustrated in \figref{fig:metadata}.

Specifically, each quadruple consists of: (i) a real source image, (ii) its tampered counterpart, (iii) the raw per-pixel difference map from which alternative labels for other $\tau$ values can be derived, and (iv) the recommended binary pixel-level label map $\mathbf{M}_{\tau}$ from a default $\tau$.
The accompanying metadata records detailed information on image generation, fidelity assessment process, and label construction.
Notably, we provide text of detailed manipulation description as presented in \appref{sec:app_text}.
Moreover, at the semantic level, beyond the semantic label, each tampered image is also accompanied by quantitative indicators such as the tampered size.

\section{Training Framework}
\label{sec:training}

\begin{figure}[h]
    \centering
    \includegraphics[width=0.8\linewidth]{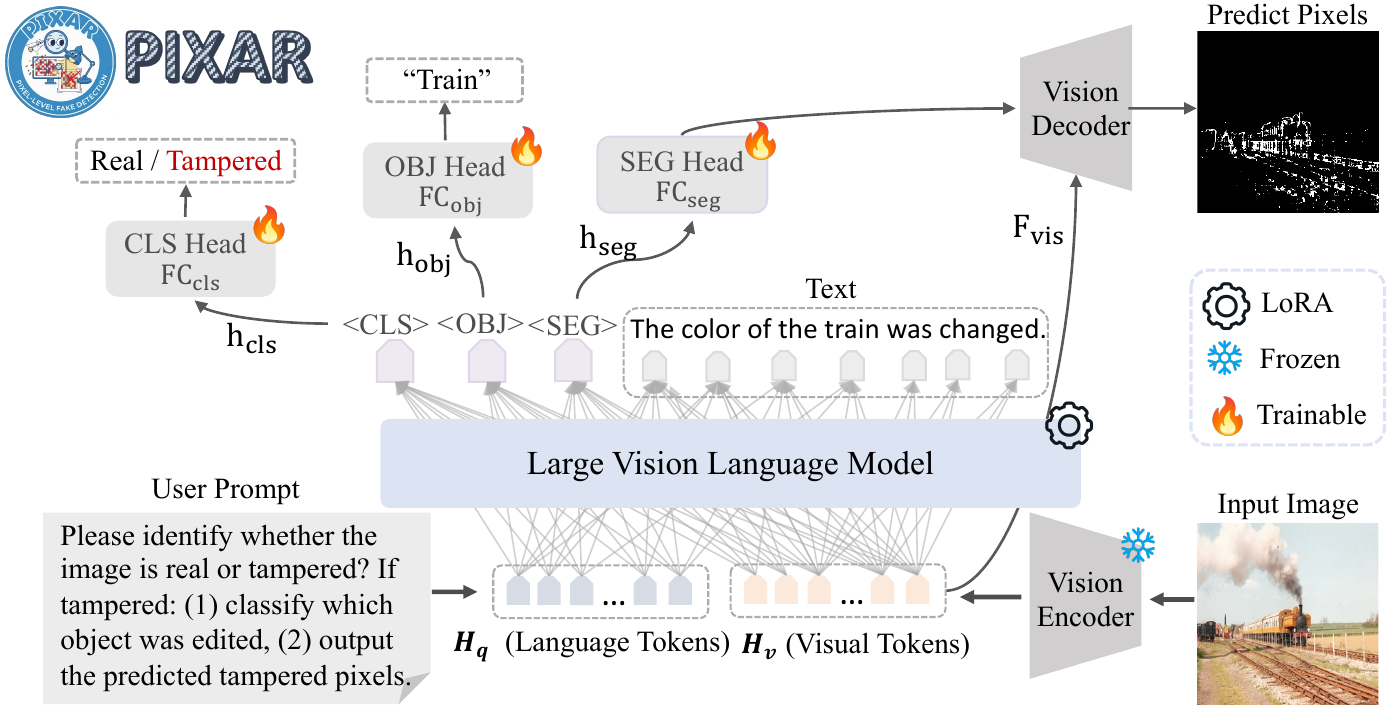}
    \caption{Overview of training framework.}
    \label{fig:method}
\end{figure}

Our tamper detector $f_\theta$ produces (i) a per-pixel tamper logit map $\mathbf{S}\in\mathbb{R}^{H\times W}$ and probabilities $\widehat{\mathbf{M}}=\sigma(\mathbf{S})$, and (ii) a multi-label semantic logit vector $\mathbf{z}\in\mathbb{R}^{|\mathcal{C}|}$ with $\hat{\mathbf{y}}=\sigma(\mathbf{z})$. Here $\sigma(\cdot)$ is the element-wise sigmoid, also a global vector for real or tamper detection. (iii) a natural language description detailing the specific tampering artifacts.
\figref{fig:method} illustrates our training framework, $\bf h$ represents the hidden feature vectors for different heads. We define the following losses:

\vspace{3pt}
\noindent{\bf Multi-label semantic loss.} Because one or multiple objects can be tampered in one image, we train with a sigmoid cross-entropy (per class):
\begin{equation}
    \setlength{\abovedisplayskip}{4pt}
    \setlength{\belowdisplayskip}{4pt}
\mathcal{L}_\text{sem}=-\frac{1}{|\mathcal{C}|}\sum_{c \in \mathcal{C}}\left[y_c \log \hat{y}_c+\left(1-y_c\right) \log \left(1-\hat{y}_c\right)\right],
\label{eq:sigmoid_cross_entropy}
\end{equation}
where $y_c\in\{0,1\}$ denotes the ground-truth for class $c$, and $\hat{y}_c$ denotes the predicted probability for tampered area semantic label.

\vspace{3pt}
\noindent{\bf Pixel-wise BCE loss.} We supervise the localization head with pixel-wise binary cross-entropy against our thresholded label $\mathbf{M}_{\tau}$:
\begin{equation}
    \setlength{\abovedisplayskip}{4pt}
    \setlength{\belowdisplayskip}{4pt}
    \begin{aligned}
    \mathcal{L}_\text{bce}&=-\frac{1}{H W} \sum_{i, j}[\mathbf{M}_\tau(i, j) \log \widehat{\mathbf{M}}_{i j} +(1-\mathbf{M}_\tau(i, j)) \log (1-\widehat{\mathbf{M}}_{i j})],
    \end{aligned}
\end{equation}
where $H$, $W$ are the height and width of an image.

\vspace{3pt}
\noindent{\bf Pixel-level DICE loss.} Additionally, to further improve the localization of tampered pixels, we compute the DICE score~\citep{azad2023loss} over connected-component masks. During training, we replace this surrogate mask with our pixel-level label $\mathbf{M}_{\tau}$ to more accurately reflect the true editing footprint:
\begin{equation}
\begin{gathered}
    \setlength{\abovedisplayskip}{4pt}
    \setlength{\belowdisplayskip}{4pt}
 \mathcal{L}_\text{dice}=1-\frac{2 \sum_{i, j} \widehat{\mathbf{M}}_{i j} \mathbf{M}_\tau(i, j)+\varepsilon}{\sum_{i, j} \widehat{\mathbf{M}}_{i j}+\sum_{i, j} \mathbf{M}_\tau(i, j)+\varepsilon}, 
\end{gathered}
\end{equation}
with a small $\varepsilon>0$ for numerical stability.

\vspace{3pt}
\noindent{\bf Global image-level detection loss.} To decide whether an image is real or tampered\footnote{We do not consider the fully synthetic category used in SIDA as it is a special case when all pixels are tampered. This case is already covered by our pixel-level detection, which is distinct from mask-based training.}, we use a global detection head driven by a special token $\langle\text{CLS}\rangle$ . Let the backbone produce the last hidden states $\mathbf{H}^{\text{hid}}\in\mathbb{R}^{N\times d}$, we extract the $\langle\text{CLS}\rangle$ representation $h_{\mathrm{cls}}=\mathbf{H}^{\mathrm{hid}}[\mathrm{CLS}] \in \mathbb{R}^d$, and feed it to a detection head $F_{\text{cls}}$ to obtain global class logits and probabilities:
\begin{equation}\small
    \setlength{\abovedisplayskip}{4pt}
    \setlength{\belowdisplayskip}{4pt}
\begin{gathered}
\mathbf{u}=F_{\mathrm{cls}}\left(h_{\mathrm{cls}}\right) \in \mathbb{R}^2, \quad \hat{\mathbf{p}}=\operatorname{softmax}(\mathbf{u}),\\
\mathcal{L}_\text{cls}=\mathcal{L}_{\text{CE}}(\hat{\mathbf{p}}, \mathbf{d}),
\end{gathered}
\end{equation}
where $\mathcal{L}_{\text{cls}}$ is the global detection loss, and $\mathbf{d}\in\{0,1\}^{2}$ is the one-hot ground truth over \{\text{real}, \text{tampered}\}.

\vspace{3pt}
\noindent{\bf Tamper description generation loss.}
To provide a human-readable explanation of the manipulation, we generate a natural language description characterizing the tampered content (e.g., ``A banana was added to the image.''). 
We use a multimodal causal language model conditioned on the input image and the textual prompt to model the autoregressive likelihood of the target token sequence $T^*=(t_1^*,\dots,t_L^*)$:
\begin{equation}
\label{eq:text-loss}
    \setlength{\abovedisplayskip}{4pt}
    \setlength{\belowdisplayskip}{4pt}
    p_\phi(T^* \mid \mathbf{I}, P) = \prod_{i=1}^{L} p_\phi\!\left(t_i^* \mid t_{<i}^*, \mathbf{I}, P\right), 
\mathcal{L}_{\text{text}} = - \sum_{i=1}^{L} \log p_\phi\!\left(t_i^* \mid t_{<i}^*, \mathbf{I}, P\right),
\end{equation}
where $\mathcal{L}_{\text{text}}$ is the standard language modeling loss and $P$ is the input prompt.

\vspace{3pt}
\noindent{\bf Final objective.} Our training objective is to minimize a weighted combination of the five losses:
\begin{equation}
    \setlength{\abovedisplayskip}{4pt}
    \setlength{\belowdisplayskip}{4pt}
\mathcal{L}_\text{total}=\lambda_\text{sem} \mathcal{L}_\text{sem}+\lambda_\text{bce} \mathcal{L}_\text{bce}+\lambda_\text{dice} \mathcal{L}_\text{dice}+\lambda_\text{text} 
\mathcal{L}_\text{text}+\lambda_\text{cls} 
\mathcal{L}_\text{cls}.
\end{equation}
where $\lambda_\text{sem}, \lambda_{\text{bce}}, \lambda_{\text{dice}}, \lambda_\text{text}, \lambda_{\text{cls}}>0$ control the trade-offs between semantic understanding and pixel-accurate localization. We follow SIDA to choose values for $\lambda_{\text{bce}}$ and $\lambda_{\text{cls}}$, and conduct ablations to determine the optimal value of $\lambda_\text{sem}$, $\lambda_{\text{dice}}$, and $\lambda_\text{text}$ in \secref{sec:exp_ablation}.
Unless stated otherwise, we use $\tau=0.05$ to form $\mathbf{M}_{\tau}$; sweeping $\tau$ at training or validation time provides a principled knob to balance micro-edit sensitivity against conservative high-precision localization.
More analysis for $\tau$ selection is provided in \secref{sec:exp_ablation}.

\section{Experiments}
\label{sec:evaluation}
Extensive experiments are performed to demonstrate the efficacy of our training framework and the redefined ground truth. We also benchmark the performance of various state-of-the-art detectors on the challenging \algopt test set to provide a rigorous comparative analysis.

\begin{table*}[!t]
    \centering
    \caption{\textbf{Pixel-level Tampered Region Localization Results and Associated Semantic Prediction Accuracy.}
    \algopt -Lite models are fine-tuned on a subset of our training data that contains masks to guide the generation of tampered images. In this setting, LISA and SIDA utilize these masks as ground-truth, whereas our model is supervised by the pixel-difference map with a threshold of $\tau = 0.05$. 
    All methods use the same backbone model, either LISA-7B or LISA-13B as specified.}
    \label{tab:comparison-sida}
    \vspace{-5pt}
    \setlength\tabcolsep{5.5pt}
    \resizebox{.86\textwidth}{!}{
    \begin{tabular}{lccccccc}
        \toprule
        Methods & \multicolumn{2}{c}{\textbf{Semantic Classification}} & \multicolumn{5}{c}{\textbf{Pixel Localization}} \\
        \cmidrule(lr){2-3} \cmidrule(lr){4-8}
        & Top-1 Acc & Top-5 Acc & Recall & F1-Score & AUC & g-IoU & IoU \\
        \midrule
        LISA-7B~\citep{lai2024lisa}   & 27.1 & 71.6 & 10.0 & 15.4&55.0&7.7& 8.3\\
        SIDA-7B~\citep{huang2025sida} & 27.1&71.9&15.0&21.1&55.0&10.7&11.8\\
        \rowcolor[HTML]{F3FBFF}
        \algopt-7B-Lite (Ours) & 28.2& 75.0& 26.4&26.1& 55.2 & 14.3 &15.0\\ \hdashline
        \myrowcolor
        \algopt-7B (Ours)   &\textbf{36.2} & \textbf{77.0} & \textbf{29.8} & \textbf{30.6} & \textbf{62.2} & \textbf{16.1} & \textbf{18.1} \\ \midrule
        LISA-13B~\citep{lai2024lisa}  & 30.6&75.1&11.4&17.3&55.6&9.0&9.5 \\
        SIDA-13B~\citep{huang2025sida}& 30.8 & 75.4 & 13.2 &19.5& 55.6&10.7&10.8 \\
        \rowcolor[HTML]{F3FBFF}
        \algopt-13B-Lite (Ours)  & 30.9 & 76.0 & 26.2 & 26.7 & 55.7 & 15.0 & 15.4\\ \hdashline
        \myrowcolor
        \algopt-13B  (Ours)      &  \textbf{37.4}& \textbf{76.0}  & \textbf{33.6}& \textbf{32.3}& \textbf{62.2}& \textbf{17.8} &  \textbf{19.3}  \\
        \bottomrule
    \end{tabular}
    }
\end{table*}

\subsection{Experimental Setup}
\label{sec:setup}
\noindent{\bf Training and Test Dataset Details.}
For both the training and test datasets, we set the threshold to $\tau = 0.05$ by default, with results of other $\tau$ discussed in \tabref{tab:diff-tau}.
To ensure a fair evaluation of detector performance, the test dataset is carefully constructed to maintain a balanced distribution across tampered classes, tampered types, and tampered sizes, resulting in $40\text{K}$ images with their pixel-level and semantic-level labels.
The detailed procedure for constructing the test set is provided in \appref{sec:app_test_data}.

\vspace{3pt}
\noindent{\bf Implementation Details.}
Leveraging SIDA~\citep{huang2025sida} and LISA~\citep{lai2024lisa}, our model is fine-tuned on \algopt to transition from coarse to precise pixel-level localization. Through joint supervision via spatial, multi-label semantic (\eqref{eq:sigmoid_cross_entropy}), and text generation (\eqref{eq:text-loss}) losses, the framework simultaneously produces refined tamper masks, categorizes affected objects, and generates descriptive explanations.

\vspace{3pt}
\noindent{\bf Pixel-level and Semantic Metrics.}
To evaluate our framework and redefined ground truth, we introduce a novel paradigm integrating pixel-level localization with semantic classification. Moving beyond conventional binary or coarse region-level metrics, this approach enables a holistic assessment of fine-grained precision and the interpretive understanding of tampered artifacts.
We assess pixel-level localization using Recall, F1-Score, and AUC to evaluate the precision of delineated tampered regions. To rigorously measure spatial overlap, we employ image-level mean IoU (g-IoU) for per-sample robustness and dataset-level IoU (IoU) for overall pixel-wise accuracy. Furthermore, semantic alignment is quantified via Top-1 and Top-5 Accuracy, evaluating the model's proficiency in categorizing manipulated objects. 
By integrating spatial precision with semantic alignment, this interpretable framework provides a holistic benchmark for fine-grained image forensics.

\subsection{Evaluation on \algopt}\label{sec:exp_evaluation}

\begin{wraptable}{r}{0.43\columnwidth}
    \vspace{-.2in}
    \caption{Performance comparison between \algopt and FakeShield.}
    \vspace{-.05in}
    \label{tab:fake-shield_compare}
    \centering
    \renewcommand{\arraystretch}{1}
    \resizebox{1\linewidth}{!}{  
        \begin{tabular}{lccccc}
            \toprule
            Methods & \multicolumn{5}{c}{\textbf{Pixel Localization}} \\
            \cmidrule(lr){2-6} 
            &Recall & F1-Score & AUC & g-IoU& IoU\\
            \midrule 
            FakeShield & 10.7&17.0&52.0&8.4&9.3 \\
            \myrowcolor
            \algopt-7B  &29.8 & 30.6 &62.2 & 16.1 & 18.1 \\
            \myrowcolor
            \algopt-13B & \textbf{33.6}&\textbf{32.3}&\textbf{62.2} & \textbf{17.8} &\textbf{19.3}\\
            \bottomrule
        \end{tabular}
    }
    \vspace{-.05in}
\end{wraptable}
\textbf{Semantic Alignment and Localization Results.}
\tabref{tab:comparison-sida} reports the detection performance on the \algopt test set. We select LISA and the SOTA method SIDA~\citep{huang2025sida} as baselines for comparison. The results show that replacing coarse masks with our pixel-level labels simultaneously improves the model’s ability to accurately localize tampered regions and its semantic consistency in predicting the manipulation context. Notably, by merely refining the supervision signal, \algopt-7B-Lite significantly outperforms SIDA-7B, elevating IoU from 6.9\% to 14.9\% and Top-1 Acc from 10.6\% to 29.5\%. This performance is further bolstered in \algopt-7B, which achieves an additional 3.2\% gain when scaled to the full training set. \tabref{tab:fake-shield_compare} further benchmarks our approach against FakeShield~\citep{xu2024fakeshield}, where \algopt demonstrates a commanding lead across all evaluated metrics. Most notably, \algopt-7B achieves a near-doubling of localization accuracy, elevating the IoU from 9.3\% to 18.1\%. Collectively, these findings corroborate the effectiveness of accurate pixel-level supervision in achieving high-precision localization and the robustness of our framework across various evaluation settings.

\begin{table*}[!t]
    \centering
    \caption{ Comparison of existing deepfake detection methods and our models evaluated on the \algopt test set.
    }
    \vspace{-5pt}
    \label{tab:evaluation_detectors}
    \resizebox{1.0\textwidth}{!}{
    \begin{tabular}{llcccccccccccc}
        \toprule
        \textbf{Detector} & \textbf{Backbone} & \multicolumn{2}{c}{\textbf{Data Distribution}} & \multicolumn{3}{c}{\textbf{Precision}} & \multicolumn{3}{c}{\textbf{Recall}}& \multicolumn{3}{c}{\textbf{F1-Score}} \\
        \cmidrule(lr){3-4} \cmidrule(lr){5-7} \cmidrule(lr){8-10} \cmidrule(lr){11-13}
         &  & GANs & Diffusion & Real & Fake & Overall  & Real & Fake & Overall & Real & Fake & Overall  \\ \midrule
        CnnSpot~\citep{wang2020cnn} & ResNet-50~\citep{he2016deep} & \textcolor{green1}{\ding{51}} & \textcolor{gray}{\ding{55}} & 15.2& 84.4 & 49.8 & 99.4 & 0.6&50.0 &26.4&1.2&13.8\\

        AntifakePrompt~\citep{chang2023antifakeprompt} & InstructBLIP~\citep{dai2023instructblip}& \textcolor{green1}{\ding{51}} & \textcolor{green1}{\ding{51}} & 4.6&84.8&44.7&0.01&99.9&50.0&0.03&91.7&45.9 \\
        
        SIDA-7B~\citep{huang2025sida} & LISA-7B~\citep{lai2024lisa}  & \textcolor{gray}{\ding{55}} & \textcolor{green1}{\ding{51}} & 15.8&86.7&51.3&88.2&13.6&50.9&26.8&23.4&25.1 \\

        SIDA-13B~\citep{huang2025sida} & LISA-13B~\citep{lai2024lisa} & \textcolor{gray}{\ding{55}} & \textcolor{green1}{\ding{51}} &  16.5& 86.3&51.4 & 62.0& 42.8 & 52.4 &26.0 &57.2 &41.6 \\
        
        \midrule
        \myrowcolor

        \algopt-7B & SIDA-7B~\citep{huang2025sida}  & \textcolor{gray}{\ding{55}} & \textcolor{green1}{\ding{51}} & 39.1&97.6&68.4&89.9&74.9&82.4&54.5&84.8&69.7\\
        \myrowcolor
        \algopt-13B & SIDA-13B~\citep{huang2025sida} & \textcolor{gray}{\ding{55}} & \textcolor{green1}{\ding{51}} & 33.6 & 98.4 & 66.0&93.4&66.7&80.1&49.5&79.5&64.5\\
        \bottomrule
    \end{tabular}}
\end{table*}

\vspace{5pt}
\noindent{\bf Binary Classification Performance.}
We evaluate a range of open-source deepfake detectors, CnnSpot~\citep{CNNDetection}, AntifakePrompt~\citep{chang2023antifakeprompt}, and SIDA-7B/13B~\citep{huang2025sida} against our \algopt models. As shown in \tabref{tab:evaluation_detectors}, our models consistently outperform all baselines, highlighting their superior generalization, precise pixel-level localization, and robust binary classification.

\begin{table*}[t]
    \centering
    \caption{Comparison of detection performance across different generative sources.}
    \label{tab:across-generative}
    \setlength\tabcolsep{5.5pt}
    \vspace{-5pt}
    \resizebox{.87\textwidth}{!}{
    \begin{tabular}{lccccccc}
        \toprule
        Methods & \multicolumn{2}{c}{\textbf{Semantic Classification}} & \multicolumn{5}{c}{\textbf{Pixel Localization}} \\
        \cmidrule(lr){2-3} \cmidrule(lr){4-8}
        & Top-1 Acc & Top-5 Acc & Recall & F1-Score & AUC & g-IoU & IoU \\
        \midrule
        GPT-Image-1.5~\citep{gptimage15} & 26.0 & 70.5 & 16.5 & 20.9 & 52.3 & 11.3 & 11.7\\
        Seedream 4.5~\citep{seedream2025seedream40} & 29.9&73.1& 27.8&26.8&57.8& 13.9&15.5  \\
        Gemini 3~\citep{gemini3} & 33.8 &75.7 & 34.5 &26.7&67.0&14.6&15.4 \\
        Gemini 2.5~\citep{gemini25} & 34.0 & 76.9 & 23.7& 29.5& 70.2 & 16.1& 17.3\\
        Flux.2~\citep{flux2} & 37.1 & 79.7 &25.7& 31.3& 56.3& 16.7& 18.6 \\
        Qwen-Image~\citep{wu2025qwen} & 52.0 & 84.0 & 56.5 & 41.6 & 76.6 & 22.6 & 26.3 \\
        \bottomrule
    \end{tabular}
    }
\end{table*}

\vspace{5pt}
\noindent{\bf Evaluation across Generative Models.}
To evaluate the cross-framework generalization of \algopt, we present its performance across diverse generative paradigms in \tabref{tab:across-generative}.
Among them, GPT-Image-1.5~\citep{gptimage15} is the most challenging case, with only 11.7\% IoU and 26.0\% Top-1 accuracy, whereas Qwen-generated images are the easiest to detect.
We attribute this gap to domain shift, since the training data is primarily generated by Qwen-based.
Despite this out-of-distribution (OOD) setting, \algopt still exhibits robust and consistent detection performance across generations from all unseen frameworks, demonstrating that it captures universal artifacts that generalize beyond specific generative backbones and architectural designs.

\subsection{Ablation Study}\label{sec:exp_ablation}

\begin{table*}[!t]
    \centering
    
    \begin{minipage}{0.45\textwidth}
        \centering
        \caption{Fixed Eval: Varying $\tau_\text{train}$, fixed $\tau_\text{eval}=0.05$.}
        \label{tab:diff-tau}
        \resizebox{1\linewidth}{!}{
        \begin{tabular}{lccccccc}
            \toprule
            & \multicolumn{2}{c}{\textbf{Semantic Class.}} & \multicolumn{5}{c}{\textbf{Pixel Localization}} \\
            \cmidrule(lr){2-3} \cmidrule(lr){4-8}
            & Top-1 & Top-5 & Recall & F1 & AUC & g-IoU & IoU \\
            \midrule
            $\tau = 0.05$ & \textbf{36.2} & \textbf{77.0} & \textbf{29.8} & \textbf{30.6} & \textbf{62.2} & \textbf{16.1} & \textbf{18.1} \\
            $\tau = 0.1$  & 35.2 & 76.1 & 15.4 & 22.4 & 60.9 & 13.1 & 12.6 \\
            $\tau = 0.2$  & 34.2 & 75.9 & 9.6  & 16.0 & 61.2 & 9.5  & 8.7  \\
            \bottomrule
        \end{tabular}
        }
    \end{minipage}
    \hfill
    \begin{minipage}{0.53\textwidth}
        \vspace{-0.119in}
        \centering
        \caption{Symmetric setting ($\tau_\text{train} = \tau_\text{eval}$).}
        \label{tab:sync-tau}
        \resizebox{1\linewidth}{!}{
        \begin{tabular}{lccccccc}
            \toprule
            & \multicolumn{2}{c}{\textbf{Semantic Class.}} & \multicolumn{5}{c}{\textbf{Pixel Localization}} \\
            \cmidrule(lr){2-3} \cmidrule(lr){4-8}
            & Top-1 & Top-5 & Recall & F1 & AUC & g-IoU & IoU \\
            \midrule
            PIXAR-7B ($\tau = 0.05$) &\textbf{36.2} & \textbf{77.0} & \textbf{29.8} & \textbf{30.6} & \textbf{62.2} & \textbf{16.1} & \textbf{18.1}  \\
            PIXAR-7B ($\tau = 0.1$) & 34.6 & 75.9 & 17.4 & 23.5 & 61.4 & 10.6 & 12.0\\
            \bottomrule
        \end{tabular}
        }
    \end{minipage}
\end{table*}

\textbf{Influence of Different $\tau$.}
To investigate the impact of training threshold $\tau$ on tampered-pixel localization, we conduct an ablation study in \tabref{tab:diff-tau} while fixing the evaluation threshold at $\tau = 0.05$. Results show that increasing the training $\tau$ consistently hampers performance. As visualized in \figref{fig:vis_tau}, a higher threshold filters out fine-grained pixel details, leaving only coarse semantic cues from the initial generation masks. This results in a less discriminative supervision signal for precise localization. To decouple the threshold's intrinsic impact from the training-test discrepancy, we evaluate a consistent setting where identical $\tau$ values are used for both phases (\tabref{tab:sync-tau}). The results show that a lower $\tau$ consistently yields superior performance, validating that $\tau=0.05$ provides a more discriminative and robust supervision signal for effective learning.

\begin{wraptable}{r}{0.5\columnwidth}
    \vspace{-.2in}
    \caption{Impact of Semantic Loss Weight.}
    \vspace{-.08in}
    \label{tab:ablation-sem}
    \centering
    \renewcommand{\arraystretch}{1}
    \resizebox{1\linewidth}{!}{  
\begin{tabular}{lccccccc}
        \toprule
        & \multicolumn{2}{c}{\textbf{Semantic Classification}} & \multicolumn{5}{c}{\textbf{Pixel Localization}} \\
        \cmidrule(lr){2-3} \cmidrule(lr){4-8}
        & Top-1 Acc & Top-5 Acc & Recall & F1-Score & AUC & g-IoU & IoU \\
        \midrule
        $\lambda_{sem} = 0.1$ & 35.1 &76.2 & 29.7&30.3& 62.2&15.9&17.9\\
        $\lambda_{sem} = 0.5$ &\textbf{36.2} & \textbf{77.0} & \textbf{29.8} & \textbf{30.6} & \textbf{62.2} & \textbf{16.1} & \textbf{18.1} \\
        $\lambda_{sem} = 1.0$ & 35.2 & 76.2&29.7 & 30.5&62.0& 16.0 & 18.0\\
        \bottomrule
    \end{tabular}
    }
    \vspace{-.2in}
\end{wraptable}

\vspace{5pt}
\noindent\textbf{Influence of $\lambda_{sem}$.}
\tabref{tab:ablation-sem} explores the impact of semantic loss weight $\lambda_{sem}$. While pixel-level localization performance remains consistently robust across different weights, we observe that both lower (0.1) and higher (1.0) values of $\lambda_{sem}$ lead to a marginal decline in Top-1 accuracy. This suggests that $\lambda_{sem}=0.5$ strikes the optimal balance, providing sufficient semantic supervision without overshadowing other task objectives. We therefore adopt $\lambda_{sem}=0.5$ to ensure maximum multi-task synergy.

\begin{wraptable}{r}{0.5\columnwidth}
    \vspace{-.2in}
    \caption{Impact of Text Loss Weight.}
    \vspace{-.1in}
    \label{tab:ablation-text}
    \centering
    \renewcommand{\arraystretch}{1}
    \resizebox{1\linewidth}{!}{  
    \begin{tabular}{lccccc}
        \toprule
        & \multicolumn{2}{c}{\textbf{Semantic Classification}} & \multicolumn{2}{c}{\textbf{Pixel Localization}} & \multicolumn{1}{c}{\textbf{Text Quality}}\\
        \cmidrule(lr){2-3} \cmidrule(lr){4-5} \cmidrule(lr){6-6} 
        & Top-1 Acc & Top-5 Acc & g-IoU & IoU & Css\\
        \midrule
        $\lambda_{text} = 2.0$ & 35.6 & 76.7 & 16.1 & 18.1 & 0.51 \\
        $\lambda_{text} = 3.0$ &\textbf{36.2}&\textbf{77.0}&\textbf{16.1}&\textbf{18.1}&\textbf{0.75} \\
        $\lambda_{text} = 4.0$ & 35.9 &76.9 & 16.1 & 18.1 & 0.75\\
        \bottomrule
    \end{tabular}
    }
    \vspace{-.2in}
\end{wraptable}

\vspace{5pt}
\noindent\textbf{Influence of $\lambda_{text}$.}
\tabref{tab:ablation-text} investigates the trade-off controlled by $\lambda_{text}$. While a lower $\lambda_{text}$ leads to suboptimal text generation, an excessively high value (e.g., $\lambda_{text}=4.0$) tends to degrade semantic accuracy. We observe that $\lambda_{text}=3.0$ strikes an optimal balance, delivering superior text quality without compromising core detection performance; thus, it is adopted as our default configuration.

\begin{wraptable}{r}{0.5\columnwidth}
    \vspace{-.2in}
    \caption{Impact of Dice Loss Weight $\lambda_{dice}$.}
    \vspace{-.05in}
    \label{tab:ablation-dice}
    \centering
    \renewcommand{\arraystretch}{1}
    \resizebox{1\linewidth}{!}{  
    \begin{tabular}{lccccccc}
        \toprule
        & \multicolumn{2}{c}{\textbf{Semantic Classification}} & \multicolumn{5}{c}{\textbf{Pixel Localization}} \\
        \cmidrule(lr){2-3} \cmidrule(lr){4-8}
        & Top-1 Acc & Top-5 Acc & Recall & F1-Score & AUC & g-IoU & IoU \\
        \midrule
        $\lambda_{dice}=0.0$ & 35.3 & 76.3 & 12.9&19.5&61.6&7.4&10.8  \\
        $\lambda_{dice}=0.5$ & 36.0 & 76.6 & 22.8 & 27.3&62.0&13.5&15.8\\
        $\lambda_{dice}=1.0$ &\textbf{36.2} & \textbf{77.0} & \textbf{29.8} & \textbf{30.6} & \textbf{62.2} & \textbf{16.1} & \textbf{18.1}\\

        \bottomrule
    \end{tabular}
    }
    \vspace{-.1in}
\end{wraptable}

\vspace{5pt}
\noindent\textbf{Influence of $\lambda_{dice}$.}  As summarized in \tabref{tab:ablation-dice}, we conduct an ablation study to evaluate the contribution of Dice loss. The results demonstrate that incorporating $\lambda_{dice}$ simultaneously enhances localization precision and semantic classification accuracy. This indicates that Dice loss provides superior spatial supervision, which refines mask boundaries and strengthens discriminative feature extraction.

\begin{figure*}[b]
    \centering
    \includegraphics[width=0.495\linewidth]{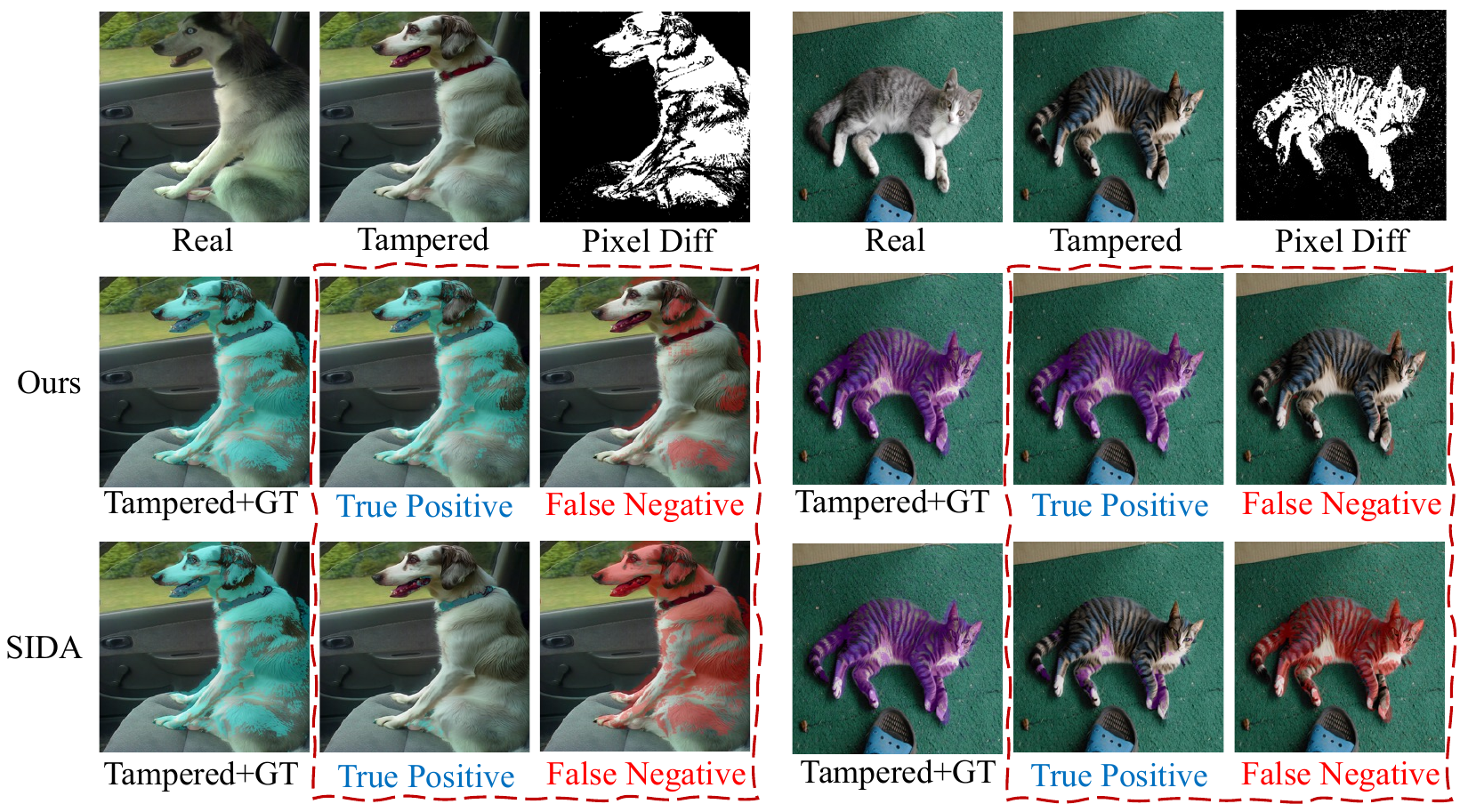}
    \includegraphics[width=0.495\linewidth]{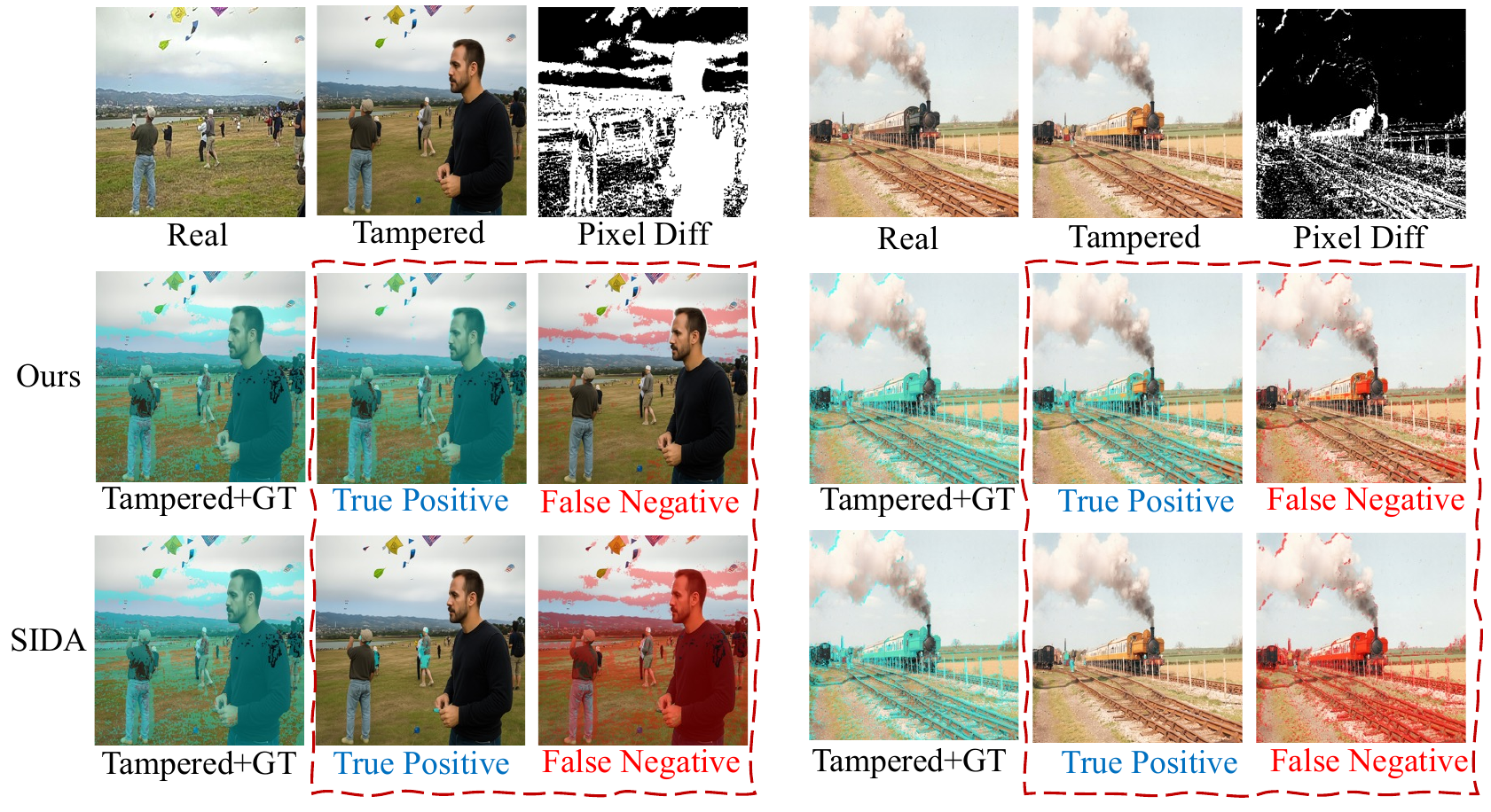}
    \vspace{-10pt}
    \caption{\textbf{Visualization comparison of prediction results between \algopt\ and SIDA~\citep{huang2025sida}.}
    The red dashed boxes show different prediction focuses compared to SIDA.}
    \label{fig:visualization_sida_ours}
    \vspace{-10pt}
\end{figure*}

\vspace{-8pt}
\subsection{Analysis}
\label{sec:exp_analysis}

\noindent {\bf Visualization.}
To visually compare the prediction results between \algopt and SIDA (after fine-tuning on our dataset), we provide a visualization in~\figref{fig:visualization_sida_ours}.
Within these red dashed boxes, we illustrate both true positives and false positives.
For true positives, the closer the predicted tampered pixels are to the actual pixel differences, the better.
False negatives refer to tampered pixels that the model fails to detect - the fewer, the better.
The figure clearly demonstrates that \textit{using masks as supervision \textbf{fails} to effectively recover the actual tampered regions}, most of the manipulated areas are missed (see the \textcolor{red1}{false negatives} of SIDA), and only a small portion of the tampered regions are correctly detected.
In contrast, our model exhibits a strong ability to accurately localize the tampered regions, with only a limited number of false negatives, further validating the effectiveness of the pixel-difference map over the ambiguous mask-based supervision in localization.

\begin{wraptable}{r}{0.5\columnwidth}
\vspace{-15pt}
    \caption{Results of user study.}
    \label{tab:userstudy}
    \centering
    \renewcommand{\arraystretch}{1}
    \resizebox{1\linewidth}{!}{  
\begin{tabular}{lcccccc}
        \toprule
        & \multicolumn{3}{c}{\textbf{Binary Classification}} & \multicolumn{3}{c}{\textbf{Localization}} \\ 
        \cmidrule(lr){2-4} \cmidrule(lr){5-7}
        & Precision & Recall & F1-Score & Recall & F1-Score & IOU \\
        \midrule
        Human & 22.2 & 55.5 & 31.0 & 17.4 & 18.8 & 10.7 \\
        \bottomrule
    \end{tabular}}
\end{wraptable}

\vspace{5pt}
\noindent {\bf User study.}
To complement our quantitative metrics, we conducted a user study to evaluate the perceptual quality of tampered images in \algopt.
A random subset of our dataset was selected for this study, consisting of $1{,}000$ images in total, including $500$ real and $500$ tampered samples.
10 participants were asked to perform two tasks: (1) classify whether the image is real or tampered, and (2) if tampered, localize the manipulated regions by drawing bounding boxes around the perceived alterations.
As shown in \tabref{tab:userstudy}, participants exhibited \textit{low performance} in both binary classification and fine-grained localization, indicating that the tampered images in our dataset achieve high visual realism.

\section{Conclusion}
\label{sec:conclusion}
In this work, we revisited VLM tampering as a pixel-grounded, meaning and language-aware task by deriving per-pixel difference maps and thresholding with $\tau$ to obtain controllable labels $\mathbf{M}_\tau$. We also release $\algopt$, a high-fidelity, large-scale benchmark of more than $420\text{K}$ image pairs built with $8$ diverse manipulations, providing original and tampered images, rich metadata, raw difference maps, recommended $\mathbf{M}_\tau$, and language descriptions for flexible supervision. We further introduced a pixel-aware training framework for localization with semantics-aware classification and natural language descriptions, and showed that state-of-the-art detectors are ill-scored under mask-only protocols, especially on micro-edits and off-mask changes, establishing a more realistic and reliable standard for fine-grained tamper detection and understanding.

\section*{Acknowledgements}

This work is supported by the United Al Saqer Group Grant.

\bibliographystyle{assets/plainnat}
\bibliography{resources/main}

\clearpage
\newpage
\beginappendix

\section{Visualization of Different $\tau$}
\label{sec:app_tau}
\begin{figure}[!h]
    \centering
    \includegraphics[width=0.7\linewidth]{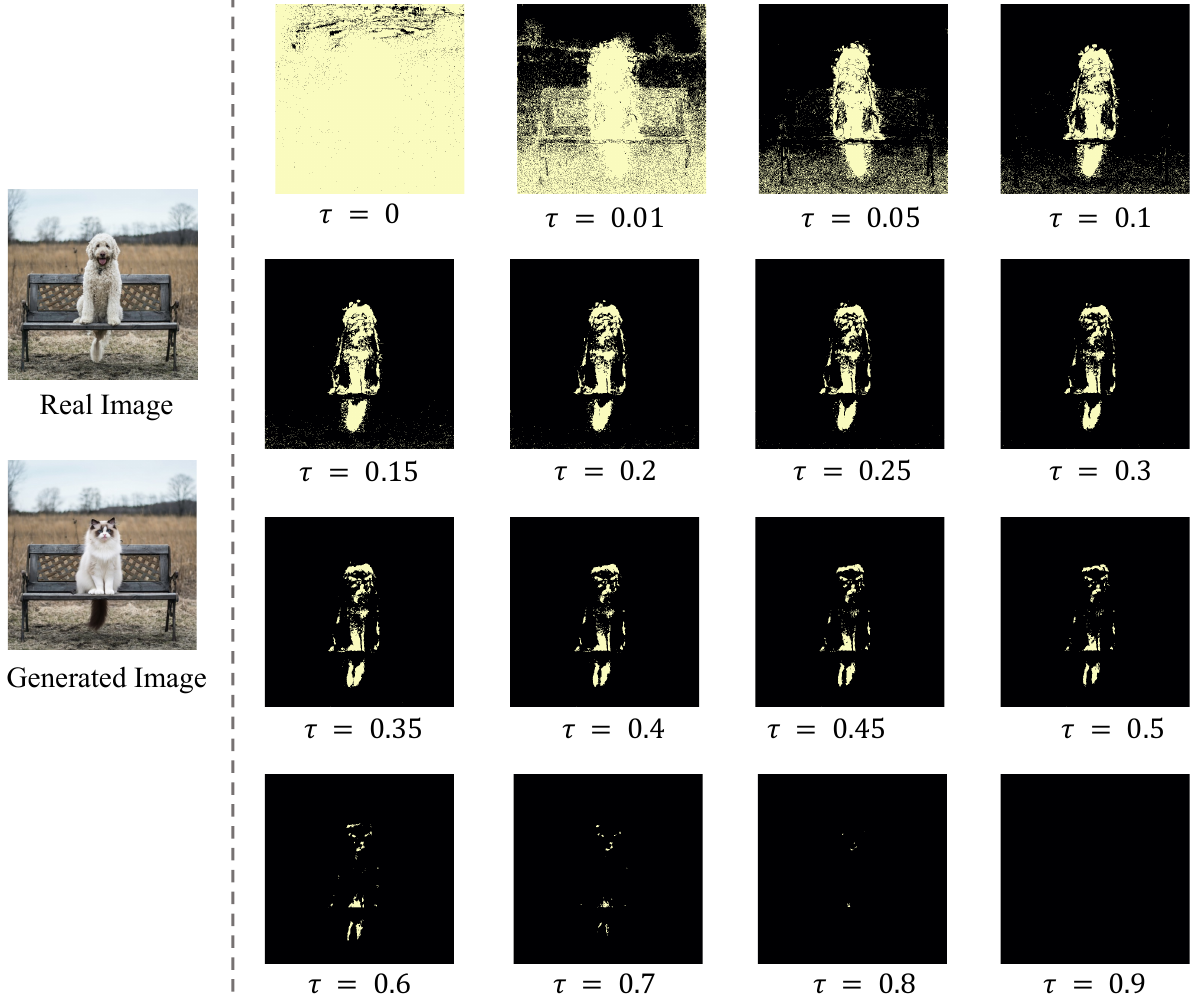}
    \caption{Visualization of pixel-level labels under different threshold values $\tau$.}
    \label{fig:app_vis_tau}
\end{figure}

Our redefined pixel label $\mathbf{M}_{\tau}$ is derived from the difference map using a tunable threshold $\tau$.
We visualize the results under different $\tau$ values in \figref{fig:app_vis_tau}.
Obviously, this threshold captures the spatial support of the edit at a controllable intensity level: a small $\tau$ emphasizes sensitivity to micro-edits, while larger $\tau$ emphasizes conservative, high-confidence changes.
This thresholded construction decouples where an edit occurs (localization) from how strongly it manifests (intensity), enabling principled sweeps over $\tau$ to select operating points that best correlate with human judgments and downstream scenario use cases.

\begin{figure}[!h]
    \centering
    \includegraphics[width=0.65\linewidth]{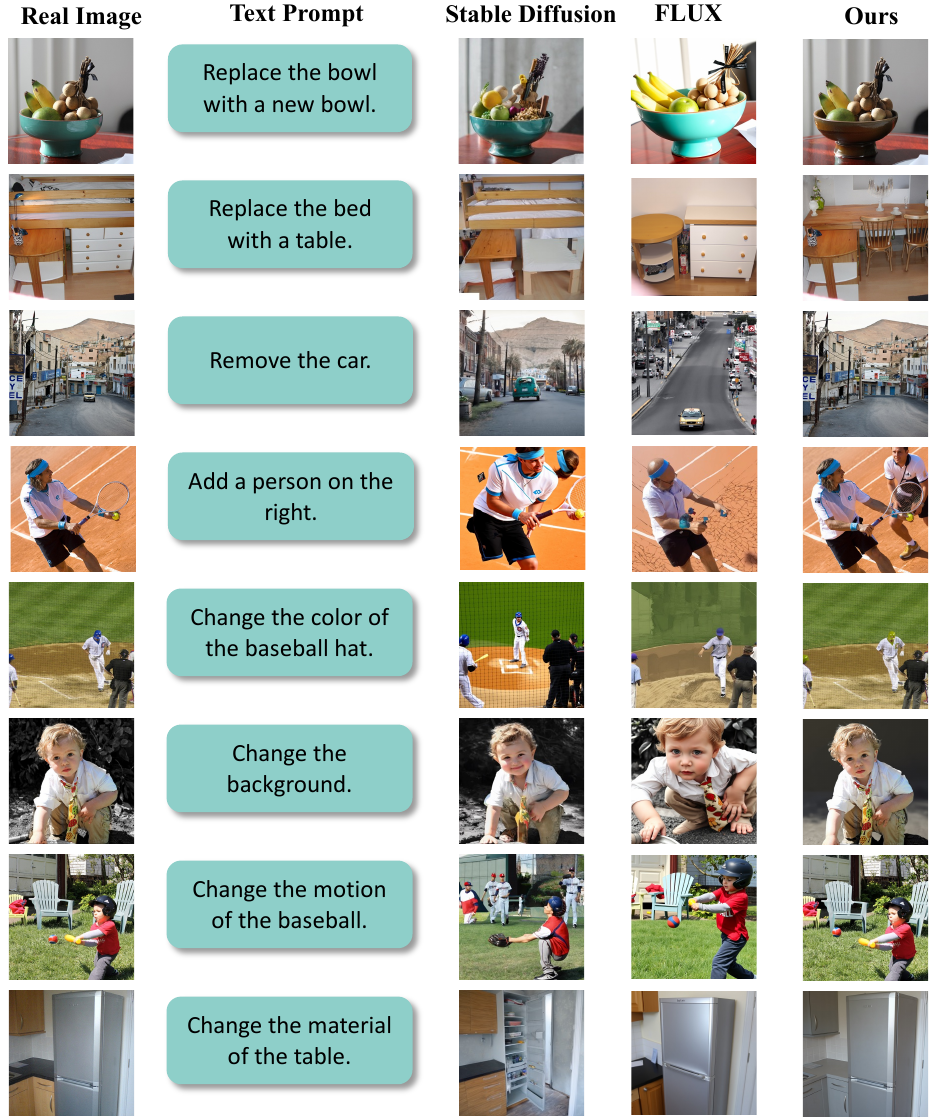}
    \caption{Visualization of tampered images generated by state-of-the-art open-source generative models.}
    \label{fig:visualization_example}
\end{figure}

\section{Additional Implementation Details for Benchmark Construction}
\label{sec:app_benchamrk}

\begin{figure*}[!t]
    \centering
    \includegraphics[width=\linewidth]{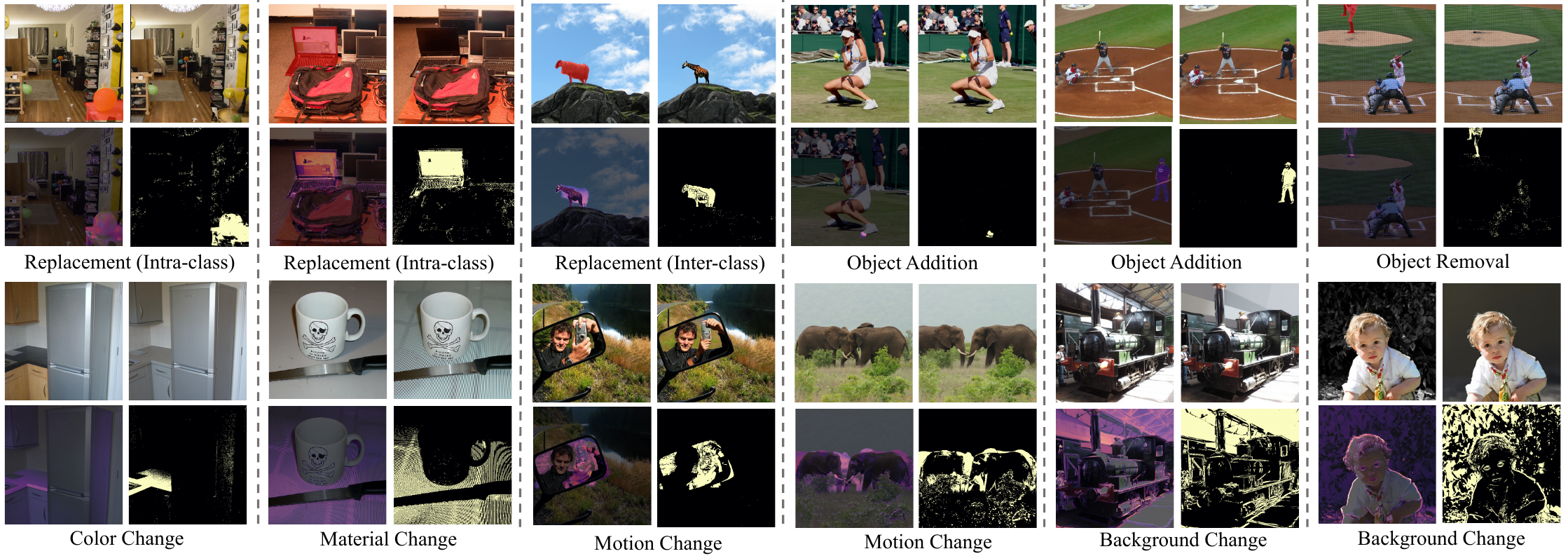}
    \caption{\textbf{Visualization of various tampering types in \algopt.}
    For each type, from top-left to bottom-right, the images show: the original image (with the tampering mask shown in red, if applicable), the generated tampered image, the pixel-difference map overlaid on the generated image, and our pixel-level label.}
    \label{fig:app_vis_benchmark}
    \vspace{-10pt}
\end{figure*}

\subsection{Image Generation}
\label{app_sec:stage_1}
\noindent{\bf Qualitative Comparison of Generative Models.}
We conduct qualitative comparisons among several state-of-the-art open-source generative models, including Flux. 2~\citep{flux2}, and Qwen-Image~\citep{wu2025qwen}.
Examples are presented in \figref{fig:visualization_example}.
Among these models, \textit{Qwen-Image VLMs consistently demonstrate higher perceptual fidelity and precise editing}, generating results that are nearly indistinguishable from real images in terms of texture realism, boundary coherence, and semantic consistency.
Therefore, we adopt Qwen-Image VLMs as the generative model for training data generation.

\begin{figure}[!t]
    \centering
    \begin{subfigure}[b]{0.46\linewidth}
        \centering
        \includegraphics[width=\linewidth]{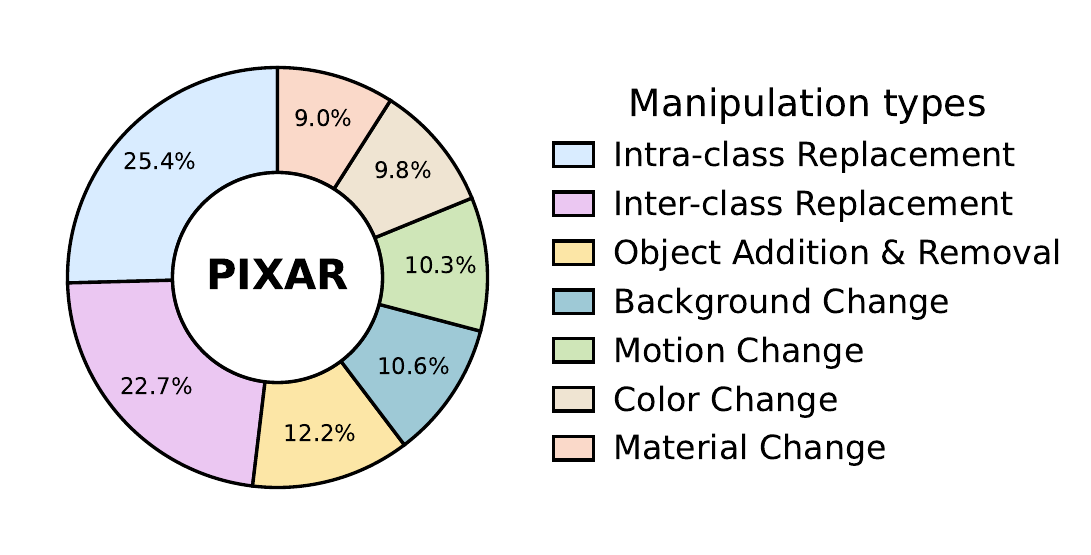}
        \caption{Tampered Type Distribution.}
        \label{fig:app_dis_tamper_type}
    \end{subfigure}
    \hfill
    \begin{subfigure}[b]{0.46\linewidth}
        \centering
        \includegraphics[width=\linewidth]{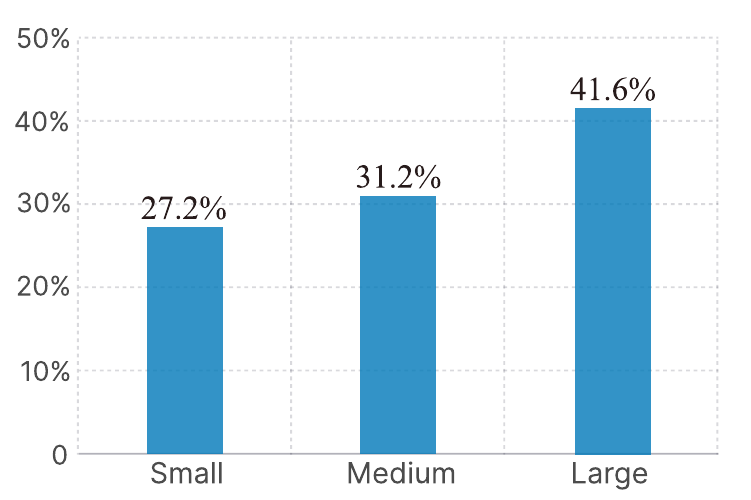}
        \caption{Tampered Size Distribution.}
        \label{fig:app_dis_tamper_size}
    \end{subfigure}
   \caption{\textbf{Analysis of \algopt Generated Images.}
    (a) shows the distribution of different manipulation types.
    (b) shows the distribution of tampered sizes.
    We set the size categorization as: \textit{small} indicates areas with fewer than \protect{$23000$} pixels, \textit{medium} indicates areas between \protect{$23000$} and \protect{$50000$} pixels, and \textit{large} indicates areas with at least \protect{$50000$} pixels.
    }
    \label{fig:app_vis_dis}
\end{figure}

\vspace{5pt}
\noindent{\bf Diverse and Practical Tampering Types.}
\label{sec:app_manipulation}
To ensure diversity, our pipeline integrates eight editing types.
Examples of different manipulation types are shown in \figref{fig:app_vis_benchmark}, and the type distribution is shown in \figref{fig:app_dis_tamper_type}.
Notably, some manipulations, such as object addition and background change, do not involve a mask during generation.
Therefore, masks are not provided for these cases in the visualization images.
Moreover, we emphasize the realistic and challenging intra-class replacement (e.g., apple $\to$ another apple), which preserves critical attributes such as object pose, scale, and contextual consistency.
These manipulations appear visually credible yet are inherently difficult to detect, thereby providing a more demanding evaluation setting.
The type distribution of the training and test set is reported in~\tabref{tab:manip_type_dist} and \figref{fig:app_test_data}~(ii).

\begin{table*}[!t]
\centering
\caption{\textbf{Dataset composition by manipulation type (including a multi-edit subset).}
Train total $N=387{,}810$, Test total $N=41{,}781$.
\textbf{Mask} indicates whether a mask is used to guide the generation of tampered images.
The three mask-conditioned training manipulations (\texttt{intra-class replacement}, \texttt{inter-class replacement}, and \texttt{object removal}; marked with $\checkmark$) form the \textsc{Lite} training subset, as referenced in~\tabref{tab:comparison-sida}.
}

\label{tab:manip_type_dist}
\vspace{-6pt}
\setlength{\tabcolsep}{5pt}
\resizebox{.9\textwidth}{!}{
\begin{tabular}{lrrc@{\hspace{14pt}}lrrc}
\toprule
\multicolumn{4}{c}{\textbf{Training set}} & \multicolumn{4}{c}{\textbf{Test set}} \\
\cmidrule(lr){1-4}\cmidrule(lr){5-8}
\textbf{Manipulation} & \textbf{\#} & \textbf{\%} & \textbf{Mask} &
\textbf{Manipulation} & \textbf{\#} & \textbf{\%} & \textbf{Mask} \\
\midrule
\texttt{intra-class replacement} & 101206 & 26.10 & $\checkmark$ &
\texttt{intra-class replacement} & 5601 & 13.41 & $\times$ \\
\texttt{inter-class replacement} &  89415 & 23.06 & $\checkmark$ &
\texttt{inter-class replacement} & 5965 & 14.27 & $\times$ \\
\texttt{object removal}          &   1059 &  0.27 & $\checkmark$ &
\texttt{object removal}          & 5694 & 13.63 & $\times$ \\
\texttt{object addition}         &  39651 & 10.22 & $\times$ &
\texttt{object addition}         & 4947 & 11.84 & $\times$ \\
\texttt{color change}            &  36687 &  9.46 & $\times$ &
\texttt{color change}            & 4456 & 10.67 & $\times$ \\
\texttt{motion change}           &  38319 &  9.88 & $\times$ &
\texttt{motion change}           & 4915 & 11.76 & $\times$ \\
\texttt{material change}         &  33981 &  8.76 & $\times$ &
\texttt{material change}         & 3993 &  9.56 & $\times$ \\
\texttt{background change}       &  39851 & 10.28 & $\times$ &
\texttt{background change}       & 4542 & 10.87 & $\times$ \\
\texttt{multi-edit}              &   7641 &  1.97 & $\times$ &
\texttt{multi-edit}              & 1668 &  3.99 & $\times$ \\
\midrule
\textbf{Total} & 387810 & 100.00 & -- &
\textbf{Total} & 41781 & 100.00 & -- \\
\bottomrule
\end{tabular}}
\vspace{-10pt}
\end{table*}

\vspace{5pt}
\noindent{\bf Diverse Tampered Sizes and Complexities.}
To evaluate across a range of difficulty levels, we control two factors: {\it tampered size} and {\it tampered complexity}.

\emph{Tampered Size.}
Detection difficulty is strongly correlated with the extent of manipulation: small-scale edits induce subtle artifacts, whereas large-scale edits typically introduce substantial semantic changes.
We define the tampered size as the {\it absolute} number of tampered pixels, and categorize the tampered size into \textit{small} ($<23,000$ pixels), \textit{medium} ($[23,000, 50,000)$), and \textit{large} ($\ge 50,000$), corresponding to approximately $7.5\%$ and $16.5\%$ of the average $640 \times 480$ image area in the COCO dataset.
Representative examples are visualized in~\figref{fig:app_vis_size}, and the overall distribution is summarized in~\figref{fig:app_dis_tamper_size}.

\emph{Multi-Object Tampering.}
Beyond the prevalent single-object editing paradigm in existing datasets (\tabref{tab:summary_datasets}), we introduce a \emph{multi-object} tampering protocol that better reflects real-world forgery pipelines, which often compose multiple heterogeneous operations through iterative editing rather than a single isolated change.
To increase compositional complexity, we construct a multi-edit subset where each image undergoes $K \in \{2, 3\}$ distinct manipulation types applied sequentially.
Concretely, we perform the first edit on the source image and feed the intermediate output as the input to the subsequent editing pass.
All hyper-parameters and prompts remain consistent with the single-object edit.
For the multiple-edit setting, the training set contains $7,641$ samples and the test set contains $1,668$ samples.
Within the test set, $459$ image pairs undergo three edits.

\begin{figure}[!t]
  \centering
  \includegraphics[width=\linewidth]{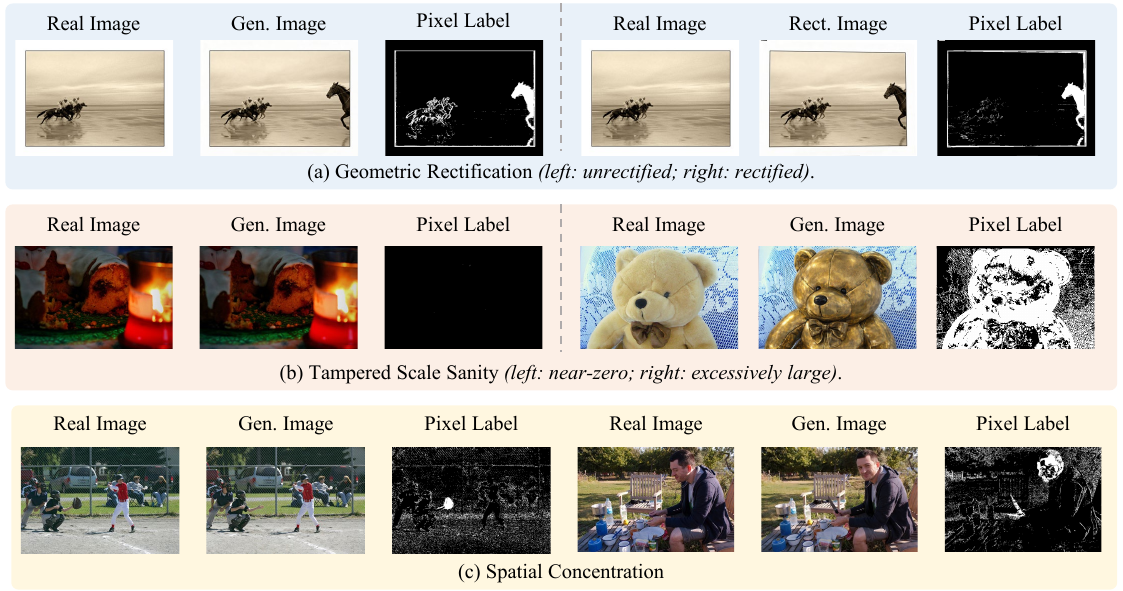}
  \vspace{-1.7em}
    \caption{\textbf{Tampering Effectiveness Checks and Label Reliability Checks.}
    (a) Geometric Rectification;
    (b) Edit Magnitude Check;
    (c) Spatial Concentration}
  \label{fig:app_Tampering_Effectiveness}
\end{figure}

\subsection{Tampering Effectiveness Checks}
\label{app_sec:stage_2}

\noindent{\bf Global Rectification.}
We perform global rectification prior to computing the pixel difference map.
Given $(I_{\text{orig}}, I_{\text{gen}})$, we first estimate a homography $H$ that maps $I_{\text{gen}}$ into the coordinate frame of $I_{\text{orig}}$ using ORB feature matching~\citep{rublee2011orb} and RANSAC~\citep{fischler1981random}.
We then warp $I_{\text{gen}}$ with $H$ to obtain the aligned image $I_{\text{gen}}^{\text{align}}$ at the same resolution as $I_{\text{orig}}$.
If homography estimation fails (e.g., insufficient matches), we fall back to using the unaligned $I_{\text{gen}}$.

Homography warping can create invalid pixels near image borders (e.g., black/undefined regions) and mild interpolation seams.
To prevent these artifacts from being interpreted as tampering signals (see~\figref{fig:app_Tampering_Effectiveness}), we detect low-intensity pixels that are connected to the image border via flood fill, dilate the detected region to cover thin seams, and replace the detected boundary pixels with the corresponding pixels from $I_{\text{orig}}$.
To avoid over-correction, we abort boundary filling if the detected boundary region exceeds a small area ratio (10\% in our implementation).

\vspace{5pt}
\noindent{\bf Edit Magnitude Checks.}
We find that the tampered area is a strong indicator of generation failure (see \figref{fig:app_Tampering_Effectiveness}~(b)).
In particular:
(i) {\it Near-zero tampering} (tampered size $\le 2{,}480$) usually corresponds to negligible modifications, yielding pixel labels with little informative signal. 
(ii) {\it Excessively large tampering} (tampered size $\ge 184{,}500$) often indicates unintended global repainting, where the pixel label includes widespread differences beyond the target semantic region and introduces substantial noise.
Therefore, we discard samples in these extreme regimes and retain only those whose tampered size falls within $[2{,}480,\,184{,}500]$.

\begin{figure}[h]
    \centering
    \includegraphics[width=.75\linewidth]{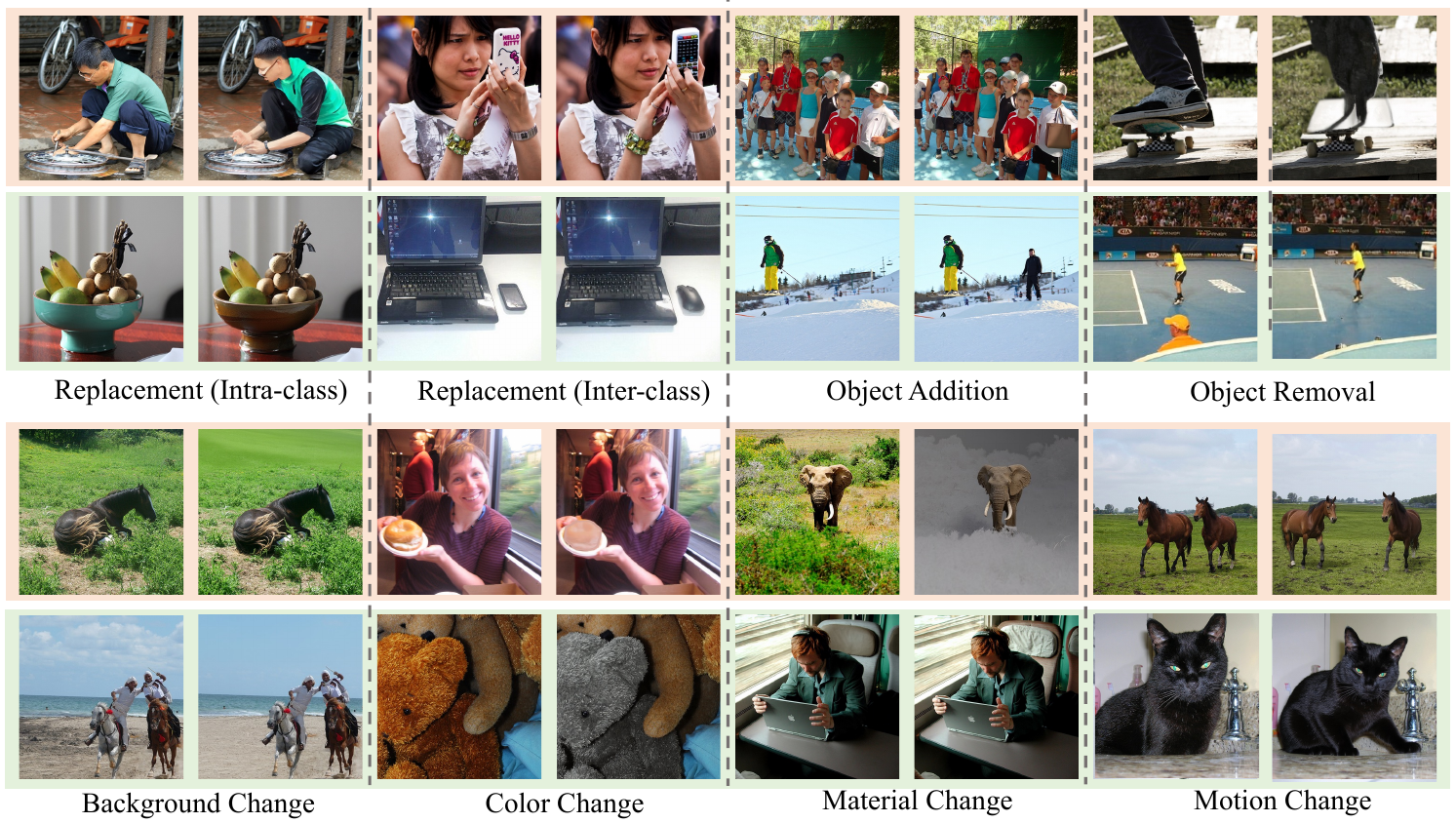}
    \caption{\textbf{Visualization of samples filtered and retained by human evaluation.}
    Images with an \textcolor{orange}{orange} background indicate filtered samples, while those with a \textcolor{green1}{green} background indicate retained images.
    }
    \label{fig:app_vis_human_filtering}
\end{figure}

\subsection{Image Fidelity Checks}
\label{app_sec:image_fidelity_check}

\noindent{\bf Human Expert Review.}
\label{sec:app_image_filter_human}
To ensure high quality of \algopt, we employ human experts to manually review the generated images and remove those that appear visually unrealistic.
Only samples that received a realism score of at least 4 out of 5 were retained.
Representative examples of filtered and retained samples in each manipulation are provided in \figref{fig:app_vis_human_filtering}.
For each manipulation, we display four images: the first column shows two real images, and the second column shows their corresponding tampered images.
Filtered samples with their corresponding real images are highlighted with an \textcolor{orange}{orange} background, while retained samples are shown with a \textcolor{green1}{green} background.

\begin{figure}[h]
    \centering
    \includegraphics[width=.8\linewidth]{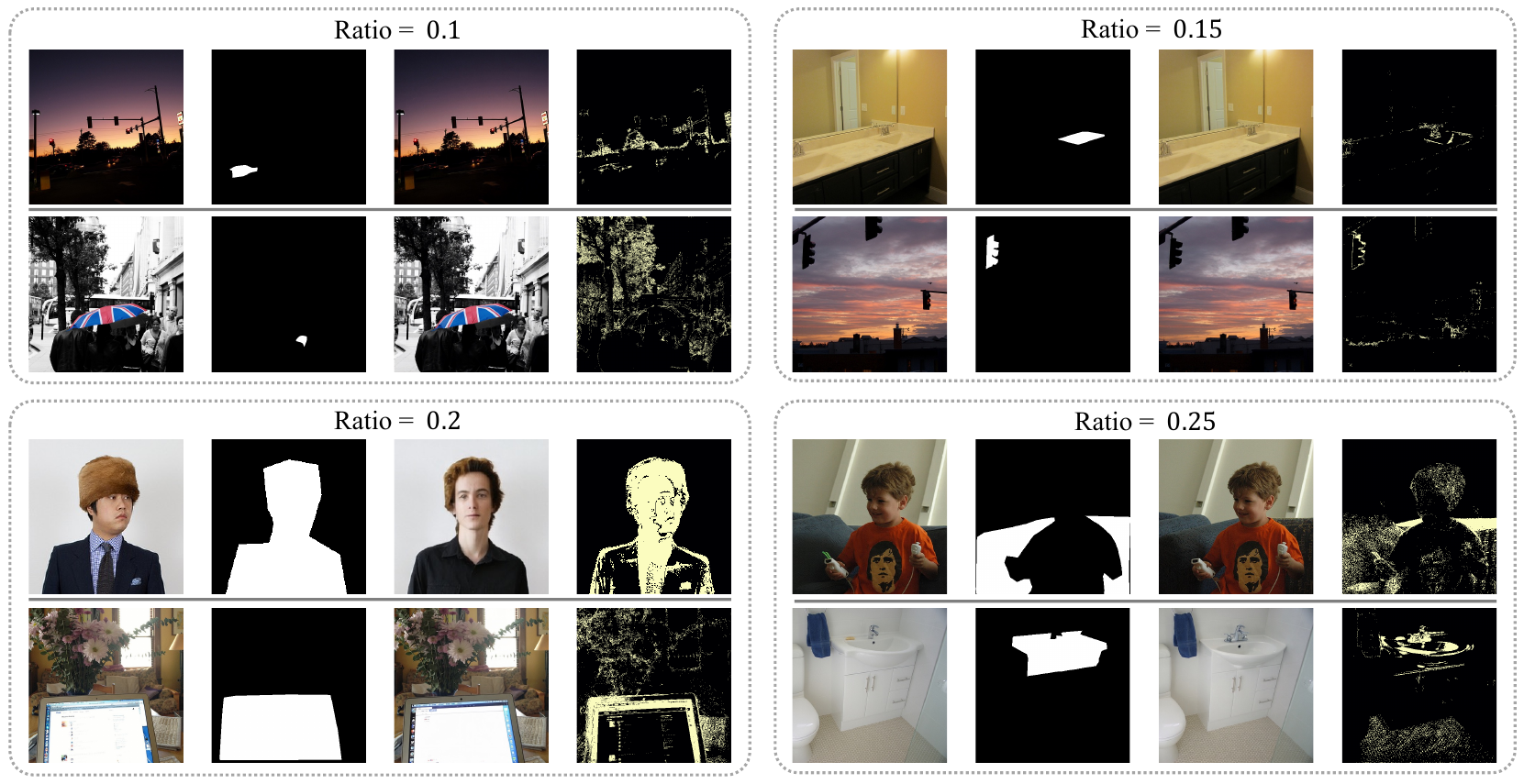}
    \caption{\textbf{Visualization of examples under different overlap ratios.}
    For each ratio, two examples are shown. For each example, from left to right: the original image, input mask, generated image, and our pixel label at $\tau=0.05$.}
    \label{fig:app_label_consistency}
\end{figure}

\subsection{Label Reliability Checks}
\label{sec:app_stage_4}
\emph{Pixel-Mask Overlap.}
In certain intra-class replacement cases (as illustrated in \figref{fig:failure_gt} of the main paper), the replaced and original objects exhibit highly similar colors and textures, making it challenging for pixel-level labels to accurately reflect true semantic tampering (i.e., the mask annotation).
This results in discrepancies between pixel-level and semantic-level annotations.

To ensure label reliability, we filter out inconsistent cases.
Specifically, we compute the overlap ratio between the tampered pixels and the input mask, and discard samples whose ratios fall below a predefined threshold, indicating substantial misalignment between pixel-level and semantic-level signals.
Representative examples across ratios are visualized in~\figref{fig:app_label_consistency}.
Obviously, as the overlap ratio increases, the consistency between pixel-level and semantic annotations improves notably.
For instance, when the ratio is around 0.10, the labels fail to capture semantic information; at 0.15, they begin to reflect partial semantic content; and when the ratio exceeds 0.20, the pixel-level labels closely align with semantic annotations.
Accordingly, we adopt a threshold of 0.2 to remove inconsistent samples, thereby maintaining the reliability of our labels.

\emph{Visualization of Spatial Concentration.}
Generation failures often yield pixel labels that are highly scattered across the image, appearing as unstructured noise rather than cohesive object boundaries (see \figref{fig:app_Tampering_Effectiveness}(c)).
While such dispersed labels can be semantically plausible, they are structurally uninformative.
Motivated by this observation, we filter out spatially dispersed pixel label maps (e.g., background speckles) using two concentration scalars computed from the binary label map $M_{\tau}\in\{0,1\}^{H\times W}$, inspired by dispersion in 2D binary patterns~\citep{taubenbock2019urbanization} and local spatial association~\citep{anselin1995local}.

\textbf{Grid coverage ratio.}
To measure the global compactness of tampered pixels, we divide the binary mask $M_{\tau}$ into a uniform $10\times10$ grid cells. 
We then define $r_{\text{grid}}$ as the smallest fraction of grid cells required to cover $80\%$ of all tampered pixels. 
A smaller $r_{\text{grid}}$ indicates that most tampered pixels are concentrated within a limited spatial region, whereas a larger value suggests a more dispersed distribution across the image.

\textbf{Local density.}
Apart from global compactness, we also introduce a local density score $r_{\text{dens}}$ to measure local spatial coherence among tampered pixels. 
Specifically, we apply a $7\times7$ mean filter to the binary mask $M_{\tau}$ and define $r_{\text{dens}}$ as the median of the resulting filtered values. 
A high $r_{\text{dens}}$ indicates that tampered pixels tend to be surrounded by other tampered pixels, suggesting a spatially contiguous or locally clustered region.

We classify each map as Concentrated or Diverse according to the fixed decision cases in~\tabref{tab:spatial-conc}, and discard Diverse samples.
All hyperparameters are fixed throughout; we evaluated multiple candidates and selected the configuration that best matches visual judgments of cohesive vs.\ speckled labels.

\begin{table}[t]
\centering
\small
\caption{\textbf{Spatial Concentration Check decision cases.} Each row specifies one case for classifying a pixel label map using $(r_{\text{grid}}, r_{\text{dens}})$ and the tie-break score $r_{\text{grid}}(1-r_{\text{dens}})$. Diverse samples are discarded in the pipeline.}
\label{tab:spatial-conc}
\resizebox{\textwidth}{!}{
\begin{tabular}{p{0.24\linewidth} p{0.24\linewidth} p{0.26\linewidth} p{0.22\linewidth}}
\toprule
$r_{\text{grid}}$ & $r_{\text{dens}}$ & $r_{\text{grid}}(1-r_{\text{dens}})$ & \textbf{Label} \\
\midrule
$r_{\text{grid}} \le 0.20$ &
\textemdash &
\textemdash &
Concentrated \\
$r_{\text{grid}} \ge 0.50$ &
\textemdash &
\textemdash &
Diverse \\
$0.20 < r_{\text{grid}} < 0.50$ &
$r_{\text{dens}} \ge 0.35$ &
\textemdash &
Concentrated \\
$0.20 < r_{\text{grid}} < 0.50$ &
$r_{\text{dens}} \le 0.25$ &
\textemdash &
Diverse \\
$0.20 < r_{\text{grid}} < 0.50$ &
$0.25 < r_{\text{dens}} < 0.35$ &
$r_{\text{grid}}(1-r_{\text{dens}}) \le 0.25$ &
Concentrated \\
$0.20 < r_{\text{grid}} < 0.50$ &
$0.25 < r_{\text{dens}} < 0.35$ &
$r_{\text{grid}}(1-r_{\text{dens}}) > 0.25$ &
Diverse \\
\bottomrule
\end{tabular}}
\end{table}

\subsection{Text Description}
\label{sec:app_text}
\begin{table*}[!t]
\centering
\caption{\textbf{Prompt templates for description construction.} We deterministically map structured metadata to single-sentence tampering description. 
For multi-edit samples, we form a compositional explanation by concatenating the corresponding single-edit descriptions in the applied order.}
\label{tab:prompt_templates}
\vspace{-6pt}
\resizebox{\textwidth}{!}{
\begin{tabular}{l p{0.52\textwidth} p{0.52\textwidth}}
\toprule
\textbf{Manipulation type} & \textbf{Template} & \textbf{Example} \\
\midrule
\texttt{background change} & \texttt{The background was changed while keeping the foreground unchanged.} & \texttt{The background was changed while keeping the foreground unchanged.} \\
\texttt{object removal} & \texttt{The \{orig\} was removed from the image.} & \texttt{The car was removed from the image.} \\
\texttt{object addition} & \texttt{A \{cat\} was added to the image.} & \texttt{A bicycle was added to the image.} \\
\texttt{intra-class repl.} & \texttt{The \{orig\} was replaced with a different-looking \{orig\}.} & \texttt{The dog was replaced with a different-looking dog.} \\
\texttt{inter-class repl.} & \texttt{The \{orig\} was replaced with a \{repl\}.} & \texttt{The chair was replaced with a sofa.} \\
\texttt{color change} & \texttt{The color of the \{cat\} was changed.} & \texttt{The color of the shirt was changed.} \\
\texttt{motion change} & \texttt{The \{cat\} was edited to show a small motion change.} & \texttt{The person was edited to show a small motion change.} \\
\texttt{material change} & \texttt{The material appearance of the \{cat\} was changed.} & \texttt{The material appearance of the table was changed.} \\
\midrule
\texttt{multi\_edit} &
\texttt{Concatenate the single-edit descriptions in the applied order.} &
\texttt{The car was removed from the image.\newline
The background was changed while keeping the foreground unchanged.} \\
\bottomrule
\end{tabular}}

\end{table*}

We follow a template-based instruction design to better leverage the fine-grained visual understanding of VLMs while keeping the conditioning signal controlled.
Specifically, we generate a single-sentence edit description leveraging each tampered image’s structured metadata (\tabref{tab:prompt_templates}).
For multi-edit samples, we additionally provide a compositional textual explanation by concatenating the corresponding single-edit descriptions in the applied order.
The instruction is intentionally minimal: it is restricted to the manipulation type, the affected category, and excludes non-semantic cues such as location, size, and image-quality descriptors.
This constrained design produces concise, semantically aligned edit descriptions that are easier for models to condition on and for humans to interpret.

\begin{table}[!t]
\centering
\caption{\textbf{Test set composition by generation model.} Total $N=41{,}781$.}
\label{tab:test_model_dist}
\vspace{-6pt}
\resizebox{0.45\linewidth}{!}{
\begin{tabular}{lrr}
\toprule
\textbf{Model} & \textbf{\#} & \textbf{\%} \\
\midrule
\texttt{Qwen-Image}~\citep{wu2025qwen} & 8185 & 19.59 \\
\texttt{GPT-Image-1.5}~\citep{gptimage15}      & 7016 & 16.79 \\
\texttt{Flux.2}~\citep{flux2}                  & 6651 & 15.92 \\
\texttt{Gemini 2.5}~\citep{gemini25}           & 6636 & 15.88 \\
\texttt{Gemini 3}~\citep{gemini3}              & 6716 & 16.07 \\
\texttt{Seedream 4.5}~\citep{seedream2025seedream40} & 6577 & 15.74 \\
\midrule
\textbf{Total} & 41781 & 100.00 \\
\bottomrule
\end{tabular}}
\end{table}

\section{Balanced Test Data Construction}
\label{sec:app_test_data}

\begin{figure*}[b!]
    \centering

    \newcommand{\coverw}{0.26\linewidth}   
    \newcommand{\coverh}{1.6em}            
    \newcommand{\covery}{-1.45em}          

    \newcommand{\xA}{0.12\linewidth}
    \newcommand{\xB}{0.44\linewidth}
    \newcommand{\xC}{0.76\linewidth}

    \newcommand{\coly}{0.6em}             

    \begin{minipage}{\linewidth}
        \centering

        \includegraphics[width=\linewidth,trim=8pt 10pt 8pt 6pt,clip]{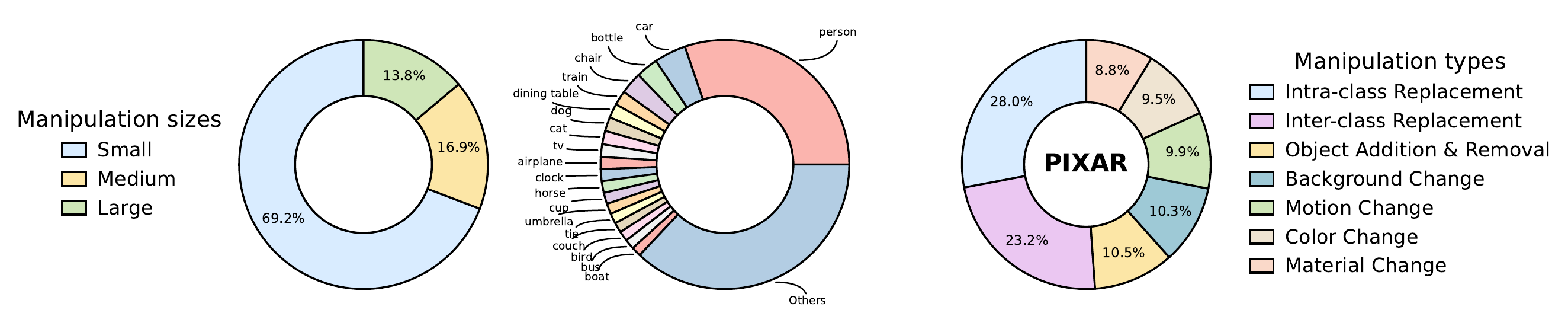}%

        \vspace{-0.2em}
        {\footnotesize\raggedright \textbf{Subfig (i)}\;Training set statistics.\par}
    \end{minipage}

    \vspace{0.15em}

    \begin{minipage}{\linewidth}
        \centering

        \includegraphics[width=\linewidth,trim=8pt 10pt 8pt 6pt,clip]{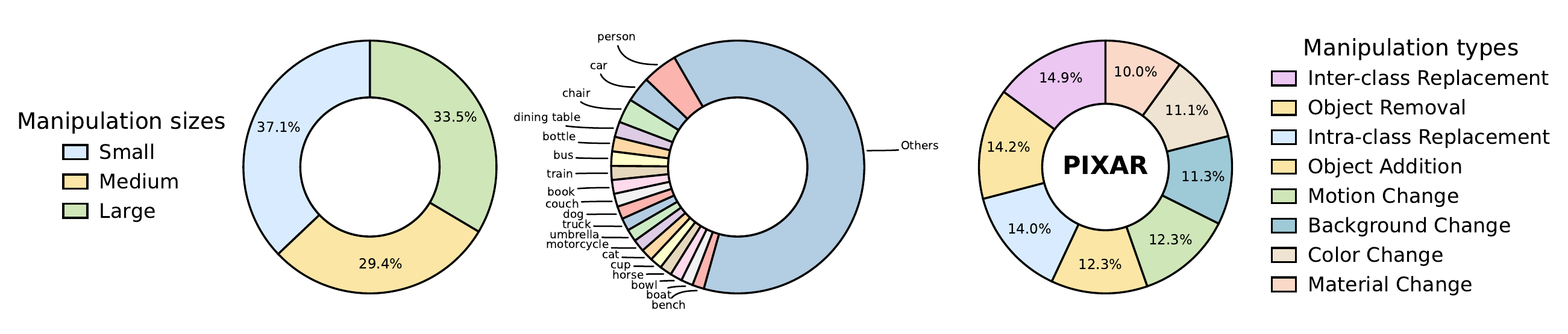}%

        \vspace{-0.2em}
        {\footnotesize\raggedright \textbf{Subfig (ii)}\;Test set statistics.\par}
    \end{minipage}

    \caption{\textbf{Dataset statistics in PIXAR.}
    The upper and lower rows visualize the training and test partitions, respectively.
    In each row, from left to right, we report the distributions over \textbf{(a) manipulation size}, \textbf{(b) manipulated object classes}, and \textbf{(c) manipulation types}.}
    \label{fig:app_test_data}
\end{figure*}

To ensure fair and comprehensive evaluation, we construct the test set with balanced distributions along three key dimensions: (i) tampered size, (ii) tampered object class, and (iii) tampering type.
Furthermore, we incorporate diverse test samples generated by $6$ state-of-the-art generative models to evaluate the generalization capabilities of detection models trained on our dataset (see~\secref{sec:exp_evaluation}).
The final test set contains over $40\text{K}$ image pairs, each annotated with both pixel-level and semantic-level labels.
All real images are sourced from the COCO validation split (\texttt{val2017})~\cite{lin2014microsoft}.
The test set further includes samples produced by multiple state-of-the-art editing models; the per-model composition of the test set is summarized in~\tabref{tab:test_model_dist}.

\noindent{\bf Tampered Size Balance.}
We also ensure varying tampered-area size among tampered samples.
We categorize the tampering scale into \textit{small} ($<23,000$ pixels), \textit{medium} ($[23,000, 50,000)$), and \textit{large} ($\ge 50,000$), corresponding to approximately $7.5\%$ and $16.5\%$ of the average $640 \times 480$ image area in the COCO dataset.

As shown in \figref{fig:app_test_data}~(ii), the final distribution is approximately balanced to \textit{Small:Medium:Large = 4:3:3},
providing a balanced composition while ensuring sufficient representation of small-scale manipulations,
which are empirically more challenging to detect~\cite{pei2024deepfake}.

\begin{table*}[!t]
\centering
\caption{\textbf{Test set class distribution (single-edit subset).} Total $N=40{,}113$. We report per-class counts and percentages.}
\label{tab:test_class_dist}
\vspace{-6pt}
\setlength{\tabcolsep}{4pt}
\resizebox{\textwidth}{!}{
\begin{tabular}{lrr@{\hspace{10pt}}lrr@{\hspace{10pt}}lrr@{\hspace{10pt}}lrr}
\toprule
\multicolumn{3}{c}{\textbf{Top 25\%}} &
\multicolumn{3}{c}{\textbf{Top 50\%}} &
\multicolumn{3}{c}{\textbf{Top 75\%}} &
\multicolumn{3}{c}{\textbf{Bottom 25\%}} \\
\cmidrule(lr){1-3}\cmidrule(lr){4-6}\cmidrule(lr){7-9}\cmidrule(lr){10-12}
\textbf{Class} & \textbf{\#} & \textbf{\%} &
\textbf{Class} & \textbf{\#} & \textbf{\%} &
\textbf{Class} & \textbf{\#} & \textbf{\%} &
\textbf{Class} & \textbf{\#} & \textbf{\%} \\
\midrule
\texttt{person} & 1812 & 4.52 & \texttt{elephant} & 586 & 1.46 & \texttt{bird} & 464 & 1.16 & \texttt{hot dog} & 370 & 0.92 \\
\texttt{car} & 1308 & 3.26 & \texttt{tv} & 578 & 1.44 & \texttt{suitcase} & 464 & 1.16 & \texttt{sports ball} & 367 & 0.91 \\
\texttt{chair} & 1240 & 3.09 & \texttt{clock} & 572 & 1.43 & \texttt{sink} & 462 & 1.15 & \texttt{tie} & 362 & 0.90 \\
\texttt{dining table} & 784 & 1.95 & \texttt{potted plant} & 568 & 1.42 & \texttt{cell phone} & 459 & 1.14 & \texttt{remote} & 348 & 0.87 \\
\texttt{bottle} & 748 & 1.86 & \texttt{sheep} & 562 & 1.40 & \texttt{fire hydrant} & 456 & 1.14 & \texttt{donut} & 345 & 0.86 \\
\texttt{bus} & 740 & 1.84 & \texttt{cow} & 555 & 1.38 & \texttt{toilet} & 451 & 1.12 & \texttt{spoon} & 341 & 0.85 \\
\texttt{train} & 717 & 1.79 & \texttt{giraffe} & 543 & 1.35 & \texttt{vase} & 441 & 1.10 & \texttt{skis} & 331 & 0.83 \\
\texttt{book} & 707 & 1.76 & \texttt{traffic light} & 542 & 1.35 & \texttt{oven} & 438 & 1.09 & \texttt{orange} & 327 & 0.82 \\
\texttt{couch} & 669 & 1.67 & \texttt{pizza} & 518 & 1.29 & \texttt{carrot} & 437 & 1.09 & \texttt{microwave} & 321 & 0.80 \\
\texttt{dog} & 657 & 1.64 & \texttt{cake} & 515 & 1.28 & \texttt{wine glass} & 425 & 1.06 & \texttt{knife} & 319 & 0.80 \\
\texttt{truck} & 645 & 1.61 & \texttt{banana} & 511 & 1.27 & \texttt{parking meter} & 423 & 1.05 & \texttt{apple} & 317 & 0.79 \\
\texttt{umbrella} & 633 & 1.58 & \texttt{stop sign} & 509 & 1.27 & \texttt{sandwich} & 422 & 1.05 & \texttt{mouse} & 306 & 0.76 \\
\texttt{motorcycle} & 629 & 1.57 & \texttt{laptop} & 500 & 1.25 & \texttt{skateboard} & 420 & 1.05 & \texttt{baseball bat} & 278 & 0.69 \\
\texttt{cat} & 627 & 1.56 & \texttt{zebra} & 499 & 1.24 & \texttt{keyboard} & 408 & 1.02 & \texttt{baseball glove} & 275 & 0.69 \\
\texttt{cup} & 621 & 1.55 & \texttt{teddy bear} & 489 & 1.22 & \texttt{kite} & 401 & 1.00 & \texttt{snowboard} & 262 & 0.65 \\
\texttt{horse} & 620 & 1.55 & \texttt{bed} & 472 & 1.18 & \texttt{fork} & 389 & 0.97 & \texttt{frisbee} & 234 & 0.58 \\
\texttt{bowl} & 616 & 1.54 & \texttt{handbag} & 468 & 1.17 & \texttt{surfboard} & 388 & 0.97 & \texttt{toothbrush} & 229 & 0.57 \\
\texttt{boat} & 611 & 1.52 & \texttt{airplane} & 468 & 1.17 & \texttt{broccoli} & 387 & 0.96 & \texttt{scissors} & 219 & 0.55 \\
\texttt{bench} & 599 & 1.49 & \texttt{bear} & 468 & 1.17 & \texttt{backpack} & 377 & 0.94 & \texttt{hair drier} & 68 & 0.17 \\
\texttt{bicycle} & 592 & 1.48 & \texttt{refrigerator} & 464 & 1.16 & \texttt{tennis racket} & 375 & 0.93 & \texttt{toaster} & 45 & 0.11 \\
\bottomrule
\end{tabular}}
\vspace{-10pt}
\end{table*}

\vspace{5pt}
\noindent{\bf Tampered Class Balance.}
COCO instance categories exhibit a naturally long-tailed distribution, with several head classes (e.g., \textit{person}, \textit{car}) dominating the dataset.
For example, in the original COCO distribution, the \textit{person} category alone accounts for approximately 30\%.
To mitigate this imbalance, we downsample overrepresented categories to achieve a more balanced distribution across classes, as illustrated in \figref{fig:app_test_data}~(ii).
This adjustment ensures that the evaluation primarily reflects the detector’s generalization ability over diverse classes rather than its performance on a few dominant ones.
We report the per-class counts and percentages of the test set in~\tabref{tab:test_class_dist} (total $N=40{,}113$).

\vspace{5pt}
\noindent{\bf Tampered Type Balance.}
The test set covers all eight manipulation types, consistent with the training set:
intra-class replacement, inter-class replacement, object removal, object addition, color change, motion change, material change, and background change.
We empirically observe that inter-class replacement and object removal exhibit lower generation success rates due to complex context blending.
For example, the generation success rate of inter-class replacement is only about one-fifth that of intra-class replacement after VLM-based scoring and is further reduced following human evaluation, as illustrated in~\secref{sec:tampering_effectiveness_check} of the main paper.
To compensate, we intentionally emphasize these low-success-rate types during data generation.
The resulting distribution across manipulation types is summarized in \figref{fig:app_test_data}~(ii).

\section{Pixel Localization Metrics}
\label{sec:app_exp_metric}

Following the evaluation protocol in our benchmark, we comprehensively assess the pixel-level localization performance of a detector using five complementary metrics: \textbf{Recall}, \textbf{F1-Score}, \textbf{AUC}, \textbf{g-IoU}, and \textbf{IoU}.  
These metrics jointly measure detection sensitivity, precision–recall trade-off, overall discrimination capability, and spatial alignment quality between predicted and ground-truth tampered regions.  
Given a pair of real and tampered images $(I_{\text{orig}}, I_{\text{gen}})$, the model produces pixel-level prediction, while the benchmark provides the pixel label $M_\tau$.  
For evaluation, we compute true positives (TP), false positives (FP), false negatives (FN), and true negatives (TN) at the pixel level.

\noindent\textbf{(1) Recall.}
Pixel-level recall quantifies the fraction of correctly detected tampered pixels among all truly tampered pixels:
\begin{equation}
\mathrm{Recall} = \frac{\mathrm{TP}}{\mathrm{TP} + \mathrm{FN}},
\end{equation}
where $\mathrm{TP}$ and $\mathrm{FN}$ denote true positives and false negatives, respectively.

\noindent\textbf{(2) F1-Score.}
Given precision $\mathrm{Prec} = \tfrac{\mathrm{TP}}{\mathrm{TP} + \mathrm{FP}}$, the F1-score provides a harmonic balance between precision and recall:
\begin{equation}
\mathrm{F1} = \frac{2 \times \mathrm{Prec} \times \mathrm{Recall}}{\mathrm{Prec} + \mathrm{Recall}} 
            = \frac{2 \times \mathrm{TP}}{2 \times \mathrm{TP} + \mathrm{FP} + \mathrm{FN}}.
\end{equation}

\noindent\textbf{(3) AUC.}
The Area Under the ROC Curve (AUC) is obtained by sweeping the decision threshold over $[0,1]$ to compute the true positive rate (TPR) and false positive rate (FPR):
\begin{equation}
\mathrm{AUC} = \int_{0}^{1} \mathrm{TPR}(\mathrm{FPR}) \, d(\mathrm{FPR}),
\end{equation}
where $\mathrm{TPR} = \tfrac{\mathrm{TP}}{\mathrm{TP} + \mathrm{FN}}$ and $\mathrm{FPR} = \tfrac{\mathrm{FP}}{\mathrm{FP} + \mathrm{TN}}$.  
This metric measures the overall discriminability of the detector independent of a specific threshold.

\noindent\textbf{(4) IoU.}
The conventional intersection-over-union (IoU) is defined as
\begin{equation}
\mathrm{IoU}
            = \frac{\mathrm{TP}}{\mathrm{TP} + \mathrm{FP} + \mathrm{FN}}.
\end{equation}
IoU directly measures the spatial overlap between predicted and ground-truth tampered pixels, providing an interpretable indicator of localization accuracy.

\noindent\textbf{(5) g-IoU.}  
The global IoU (g-IoU) in our implementation is the mean IoU across all tampered samples:
\begin{equation}
\mathrm{g\text{-}IoU}
= \frac{1}{N}\sum_{i=1}^{N}
  \frac{|\widehat{\mathbf{M}}_i \cap \mathbf{M}_{\tau,i}|}{|\widehat{\mathbf{M}}_i \cup \mathbf{M}_{\tau,i}| + \varepsilon},
\end{equation}
where $\widehat{\mathbf{M}}_i$ and $\mathbf{M}_{\tau,i}$ denote the predicted and pixel labels of sample $i$, and $\varepsilon$ is a small constant to avoid division by zero.

\end{document}